%% file: acl_latex.tex
\documentclass[11pt]{article}

\usepackage[preprint]{acl}
\usepackage{booktabs}
\usepackage{times}
\usepackage{latexsym}
\usepackage{etoolbox}
\usepackage{placeins}
\usepackage{subcaption}
\usepackage{xcolor}
\usepackage{hyperref}
\usepackage{float}

\usepackage[T1]{fontenc}

\usepackage[utf8]{inputenc}
\usepackage{multirow}
\usepackage{microtype}
\usepackage{enumitem}
\usepackage{listings}
\usepackage[most]{tcolorbox}
\newtcolorbox{promptbox}[1]{
  colback=white,
  colframe=black!70,
  coltitle=white,
  colbacktitle=black!70,
  title=#1,
  fonttitle=\bfseries,
  arc=4pt,
  boxrule=0.8pt,
  left=6pt,
  right=6pt,
  top=6pt,
  bottom=6pt,
  fontupper=\rmfamily\small
}

\usepackage{booktabs,tabularx,array}
\newcolumntype{Y}{>{\centering\arraybackslash}X}
\renewcommand{\arraystretch}{1.12}
\usepackage{tabularx}
\usepackage{array}
\usepackage{makecell}
\usepackage{xcolor}
\usepackage{booktabs}
\usepackage{amssymb}
\usepackage{inconsolata}
\usepackage{ragged2e}
\usepackage{graphicx}
\usepackage{pifont}
\usepackage{listings}
\usepackage{xcolor}
\title{VNU-Bench: A Benchmarking Dataset for Multi-Source Multimodal News Video Understanding}

\author{
Zibo Liu \quad
Muyang Li \quad
Zhe Jiang \quad
Shigang Chen
\\[0.5em]
University of Florida, Gainesville, FL, USA
\\
\texttt{\{ziboliu,muyang.li,zhe.jiang,sgchen\}@ufl.edu}
}

\begin{document}
\maketitle


\begin{abstract}
News videos are carefully edited multimodal narratives that combine narration, visuals, and external quotations into coherent storylines. In recent years, there have been significant advances in evaluating multimodal large language models (MLLMs) for news video understanding. However, existing benchmarks largely focus on single-source, intra-video reasoning, where each report is processed in isolation. In contrast, real-world news consumption is inherently multi-sourced: the same event is reported by different outlets with complementary details, distinct narrative choices, and sometimes conflicting claims that unfold over time. Robust news understanding, therefore, requires models to compare perspectives from different sources, align multimodal evidence across sources, and synthesize multi-source information. 
To fill this gap, we introduce VNU-Bench, the first benchmark for multi-source, cross-video understanding in the news domain. 
We design a set of new question types that are unique in testing models' ability of understanding multi-source multimodal news from a variety of different angles.
We design a novel hybrid human-model QA generation process that addresses the issues of scalability and quality control in building a large dataset for cross-source news understanding. The dataset comprises 429 news groups, 1,405 videos, and 2,501 high-quality questions. Comprehensive evaluation of both closed- and open-source multimodal models shows that VNU-Bench poses substantial challenges for current MLLMs.
\end{abstract}

\input{sections/introduction}

\input{sections/method}

\input{sections/experiment}
\input{sections/related_works}
\input{sections/conclusion}
\input{sections/limitation}
\bibliography{custom}

\appendix
\input{sections/appendix}

\end{document}

%% file: sections/introduction.tex
\section{Introduction}
Understanding news videos, which combine dynamic visual evidence with rich linguistic narration, has become an increasingly important subject in multimodal large language model (MLLM) research. Some existing news benchmarks~\citep{liu2021visual,luo2021newsclippings} are text-based with images, focusing on tasks of news summarization, and captioning, whereas Newsnet~\citep{wu2023newsnet} is video-based, exploring video news segmentation. Others~\citep{videoqa,gupta2022newskvqa,jahagirdar2023watching,chou2024multi}
evaluate news understanding capability through question answering, captioning, retrieval. Despite a rich body of prior work, \emph{the above benchmarks are all limited to single-sourced news reasoning.} For news videos, models interpret each  video in isolation.

In real-world journalism, stories are covered by many news outlets, 
where each report may include only partial aspects of a story, about persons involved, events, locations, timeline, organizations, etc., which must then be pieced together.
Multiple reports on the same story typically offer distinct narrative angles, visuals and commentaries, with complementary details and evolving event updates. Just as humans seek reportings about the same event from different news outlets, {\it models will understand the news events better and provide users with high-quality query results if they have superior capability of multi-sourced news reasoning.}  

Multi-sourced news video understanding poses significant challenges that require models to compare perspectives from different sources, align multimodal evidence across sources, integrate partial information from all sources, detect cross-source conflicts among claims and evidences, and synthesize a comprehensive, coherent story. These are not embodied in generic cross-video benchmarks, such as CVBench~\citep{zhu2025cvbench}, which evaluates cross-video object-event association, and CrossVideoQA~\citep{meng2025videoforest}, which evaluates person-centric reasoning by aggregating contextual cues across multiple clips. 
Also related are \cite{yao2023end,cao2025averimatec,qi2023fakesv,zeng2025understand}, which are limited to the specific task of detecting fake news or verifying truthfulness in a given text/image/video by exploring external text/image/video information. They are too narrow for a general-purpose news understanding benchmark.

With this paper, we create the first multi-sourced video news understanding benchmark (VNU-Bench), use it to evaluate a number of popular closed- or open-source models, and provide some initial insight into the ability of the state-of-the-art models in understanding multi-sourced news and answering complex questions. Our contributions are two-fold:

First, we design a set of new question types that are unique in testing models' ability of understanding multi-source multimodal news from a variety of different angles. It provides a concrete basis for future study on this subject. We then design a novel hybrid human-model QA generation process that addresses the issues of scalability and quality control in building a large dataset for cross-source news video understanding (which is much harder and time-consuming than the single-source case).  

Second, we perform comprehensive experiments that evaluate three selected closed-source MLLMs and twelve open-source MLLMs based on the new VNU-Bench dataset. The results provide some initial insight to the ability of the state-of-the-art models: Among other findings, the current MLLMs are limited in their ability of complex multi-source news video understanding, particularly on integrating information from multiple news videos (e.g., cross-source evidence-claim conflict detection); closed-source models outperform open-source models, and larger and latter models have an edge, as is predictable; models perform comparably in culture \& art news, while diverging significantly in world news.

%% file: sections/method.tex
\input{tables_figures_tex/tasks}
\section{Dataset Construction}

\input{tables_figures_tex/pipeline}
\input{tables_figures_tex/QA_example1}

\subsection{Types of Questions}
Our VNU-bench dataset contains a number of news groups. Each group contains multiple news videos from different sources (news outlets) that cover the same story/event.
Then, a number of questions are generated from each news group. These questions are intended for testing various LLM models on their understanding of multi-source multimodal news. Table \ref{tab:vnu_tasks} defines different types of news questions, Type 1 through Type 10, which are organized into two categories: (1) multi-source news comparison and (2) cross-source news integration. The multi-source news comparison category contains Types 1-5 questions, which test a model's understanding of the main claims, the details of events, the visuals, the narrative angles, and the presentations of different reportings, with comparison and contrast amongst different sources. The cross-source news integration category contains Types 6-10 questions, which test a model's ability to integrate across reportings over their audio (textual)-video evidences,  detect cross-source evidence-claim conflicts, establish the temporal order among the reportings, integrate their details into a comprehensive and coherent story, and make a cross-source summary. They are multiple-choice questions. Figure~\ref{fig:samples} gives four selected sample questions. Additional questions of all types are provided in Appendix~\ref{apd:sample_qa}.

Importantly, the question taxonomy about VNU-Bench in Table~1 is fundamentally different from that of all existing news and video benchmarks.
The prior datasets typically formulate questions over a \emph{single news report}, focusing on intra-video perception, event retrieval, or local multimodal alignment.
In contrast, all question types in VNU-Bench are explicitly designed to \emph{span multiple news videos from different outlets}, requiring models to compare claims, reconcile heterogeneous evidences, track temporal evolution, and synthesize coherent cross-source narratives.
As a result, these questions cannot be answered by reasoning over any single video in isolation.

The general process of constructing news QAs (questions and answers) is shown in Figure~\ref{fig:pipeline}. It contains steps of news group collection, QA drafting, QA quality control, and human checking, which will be explained below. 


\subsection{News Group Collection} \label{sec:col}

To ensure the diversity of VNU-Bench, we carefully curated news videos spanning a wide range of domains, including \emph{World}, \emph{US \& Canada}, \emph{Business \& Economy}, \emph{Science \& Technology}, \emph{Climate \& Environment}, \emph{Health}, and \emph{Culture \& Arts}. 
First, we created the searching events of YouTube, where each event was constructed to include a specific named entity (e.g., corporations, public figures, or geographic locations), a concrete action  (e.g., lawsuit, launch, verdict, outbreak), and a context-defining keyword. 
Then a group of volunteers reviewed and only kept the related videos. Finally, we selected multiple videos with transcripts to download, which form a news group (consisting of reports of the same event by different news outlets).

For each video, we ranked its frames based on their captions' aggregate relevance to the 10 question types. This allowed us to select informational frames that are most relevant to the target questions. The details of the ranking approach are provided in Appendix \ref{apd:frame_selection}.

\subsection{Drafting QAs}

We adopted a hybrid human-model QA generation process to draft the initial set of multi-sourced multimodal news QAs, which would then be subject to further steps of selection and quality control. 

To begin with, we manually selected 30 news groups from different domains and created 150 QAs, each taking us about 45 mins on average to view the videos, read the transcripts, analyze the transcripts and the selected frames across multiple sources in each news group, match the details in their evidences and claims, compare and contrast, come up with the question description and the answer options, and revise them in iterations. 

To scale the number of draft QAs to thousands, we painstakingly abstracted our learned knowledge in multi-sourced, multimodal QA creation into a detailed framework including: (1) 10 taskhead prompts, one for each question type, describing the goals, the core concepts, what each type-specific question must focus on, what it must avoid, the structural requirements, the source usage requirements, the guidelines about how to use transcripts and frames, and high-level sample questions; (2) a set of common rules for all question types, regarding sources, reasoning, answer options, style, anti-repetition and output format, etc. Their details can be found in the Appendix \ref{apd:qa_gen}. We fed the above framework to GPT 5 for additional draft QA generation. 



\subsection{QA Quality Control}

After generating the first-round draft questions, we performed quality control through a multi-agent assessment and filtering framework to rate the quality of the questions and removed the ones with relatively low quality. This stage was designed to enforce genuine multi-source reasoning, eliminate ambiguous questions, and assess the difficulty levels of the questions, 



First, to ensure genuine multi-source dependency of the QAs, we performed a single-source solvability test: Under a restricted input setting where only one video source was provided at a time, each question was evaluated by MLLMs-as-judge \cite{li2024llms},  where we used a set of heterogeneous multimodal large language models (MLLMs), including Gemini-2.5-pro, GPT-4o, and Qwen3-VL-30B, in order to mitigate model-specific biases~\cite{ye2025justice}. Note that these models will not be used in our model evaluation of the next section. If they could answer the question with high confidence using any single source alone, the question was removed.  This step ensures that the remaining questions cannot be answered through isolated cues from individual videos, but instead require integrating the information from multiple sources. The details of the single-source solvability test is provided in Appendix \ref{apd:single-solve}.


Second, to remove QAs with multiple correct answers, we used MLLMs-as-judge to perform ambiguity analysis on each QA. For each question, among its answer options (A through D), we pairwise-compared the correct option given by the draft, termed as the {\it choice option}, with any other option, termed as the {\it alternative option}. They were prompted to the models for whether an alternative option could also be considered correct in the context of the given news group. The models would each assign a severity score for each compared pair on semantic equivalence or ambiguity. The final severity score for a QA is defined as the maximum severity among all choice-alternative option pairs. QAs of high severity --- indicating that at least one alternative option could largely be considered as correct —-- were flagged as ambiguous QAs, which were then removed. This step ensures that each remaining QA admits a single, well-defined correct answer. More details are provided in Appendix \ref{apd:ambiguity}


Finally, we used MLLMs-as-judge to evaluate each remaining QA. Each model chose an answer and output a confidence score, together with its evidence and rational behind its choice. These outputs provided an empirical proxy for estimating question difficulty. Questions that were answered correctly by all agents with consistently high confidence were identified as overly easy questions and were removed.  More details are provided in Appendix \ref{sec:eval}.


Through the above quality control, 2240 QAs were removed from 6374 draft QAs, while 4134 QAs were kept. Specifically, 828 QAs were removed by the single-source solvability test; 637 QAs were removed by ambiguity analysis; and 775 QAs were removed by difficulty check.

\subsection{Human Checking}
As a final safeguard, we took an additional round of human verification to ensure the quality of the remaining QAs.
This verification process was conducted by a group of reviewers, all of whom were provided with a detailed verification criteria sheet (See Appendix~\ref{app:human_verification} for details.). During review, reviewers were given access to the full video transcripts with timestamps, and were required to watch all videos within a news group before evaluating the corresponding questions and answers.
With the verification score sheet, reviewers focus on aspects that MLLM judges may not reliably capture, and assign ratings across multiple dimensions, including answer correctness, language naturalness, and task compliance. Their assessment would be integrated to make final decisions. a QA is discarded if any evaluation dimension is rated as fail. Beyond verification, human reviewers also rebalanced the distribution of correct answer options  to mitigate positional bias (e.g., over-representation of certain option letters). This final human  verification step ensured that the resulting dataset consisted of high-quality, challenging, and well-balanced questions that genuinely required comprehensive multi-video understanding. In human checking, we initially have 4134 QAs from quality control, then we removed $40\%$ and finally retain 2501 QAs. 

To quantify the consistency of human verification, we randomly sampled 10\% of the QAs from 4134 QAs obtained from quality control and asked three volunteers to independently review each QA based on the same verification criteria.
We measured consistency using pairwise agreement, which captured the average proportion of exact label matches across all reviewer pairs \cite{landis1977measurement}, and additionally reported Fleiss’~$\kappa$ \cite{fleiss1971measuring} to account for agreement expected by chance among multiple reviewers.
Across axes, pairwise agreement reaches 0.75 for Correctness, 0.69 for Naturalness, and 0.82 for Task Compliance, while the overall Fleiss’~$\kappa$ is 0.62.
Given the multi-dimensional nature of our verification rubric, $\kappa$ is known to be conservative.
Overall, these results indicate that the reviewers were largely consistent under the same verification criteria. More details are in Appendix \ref{apd:iaa_metrics}.


%% file: tables_figures_tex/tasks.tex
\begin{table*}[h]
\centering
\small
\caption{\textbf{Type taxonomy of \textsc{VNU-Bench}.}
The benchmark organizes ten question types into two conceptual categories:
\textbf{(1) Multi-source News Comparison} — comparing claims, event details, visuals, narrative angles, and multimodal presentations across outlets; and
\textbf{(2) Cross-source News Integration} — integrating multi-source evidences, detecting conflicts, tracing temporal development, reconstructing story narratives, and summarizing across outlets.}
\begin{tabularx}{\textwidth}{l l X}
\toprule
\textbf{Category} & \textbf{Type ID \& Name} & \textbf{Focus} \\
\midrule
\multicolumn{3}{l}{\textbf{(1) Multi-source News Comparison}}\\

& T1. Main Claim Comparison 
& \textit{What they claim} — Compare the main conclusions different outlets draw about the same story. \\

& T2. Event Comparison
& \textit{What details they report about the events} — Compare the details of the events across sources. \\

& T3. Visual Comparison
& \textit{What they show} — Compare visual content and how it supports each outlet’s narrative. \\

& T4. Narrative Angle Comparison
& \textit{How they frame it} — Compare interpretive lenses, tone, emphasis, and perspectives used by outlets. \\

& T5. Multimodal Presentation Comparison
& \textit{How they present it} — Compare how text/audio and video evidences are integrated in each reporting. \\

\addlinespace[4pt]
\multicolumn{3}{l}{\textbf{(2) Cross-source News Integration}}\\
\addlinespace[2pt]

& T6. Cross-source Cross-modal Evidence Integration
& \textit{How to combine evidences} — Integrate text and visual evidences across outlets into a coherent reasoning chain, resolving conflicts. \\

& T7. Cross-source Evidence–Claim Conflict Detection
& \textit{Where they conflict} — Identify when evidence from one outlet contradicts claim made by another outlet. \\

& T8. Cross-source Temporal Development
& \textit{What happens first and what next} — Establish temporal ordering of event details reported by multiple outlets. \\

& T9. Cross-source Narrative Reconstruction
& \textit{How the story unfolds} — Construct a unified and coherent narrative from fragmented multi-source reports. \\

& T10. Multi-source Summary
& \textit{How to summarize} — Summarize shared points, disagreements, and overall situation across sources. \\

\bottomrule
\end{tabularx}
\label{tab:vnu_tasks}
\end{table*}

%% file: tables_figures_tex/pipeline.tex
\begin{figure*}[h]  
    \centering
    \includegraphics[width=0.75\paperwidth]{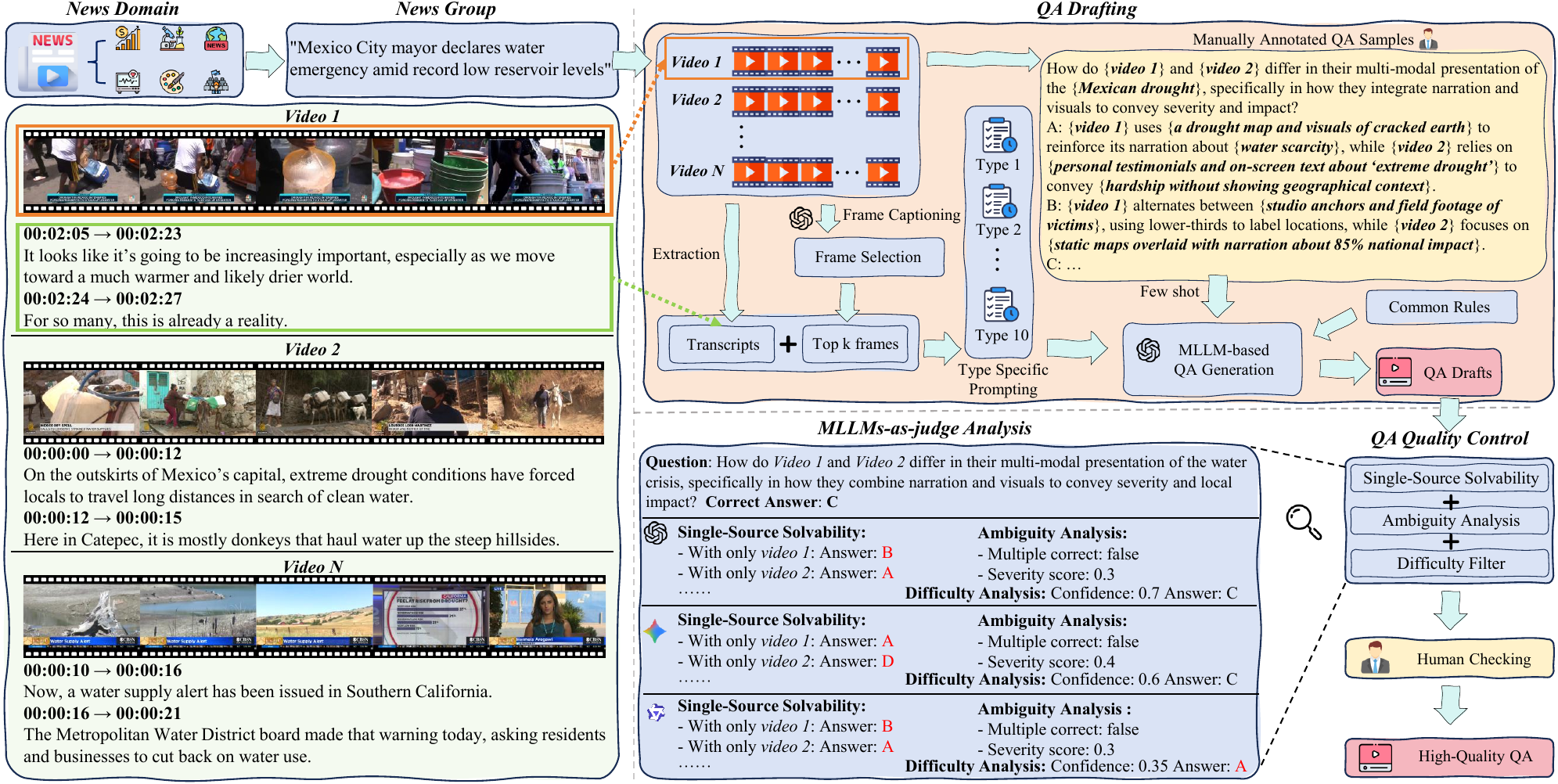}
    \caption{Overview of the VNUBench dataset construction pipeline. News video groups are collected from diverse domains, with a small subset manually annotated to serve as few-shot exemplars. Visual frames and transcripts are extracted from the videos and integrated with type-specific prompting, common rules prompting and few-shot examples to synthesize initial QA drafts via MLLMs. Subsequently, the drafts undergo a strict quality control process assisted by an MLLM-as-judge mechanism, which evaluates candidates based on single-source solvability, ambiguity analysis, and difficulty analysis to eliminate trivial samples. Finally, the filtered questions are carefully checked by voluteers to ensure the high quality of the final dataset.}
    \label{fig:pipeline}
\end{figure*}

%% file: tables_figures_tex/QA_example1.tex
\begin{figure*}[h]  
    \centering
    \includegraphics[width=0.75\paperwidth]{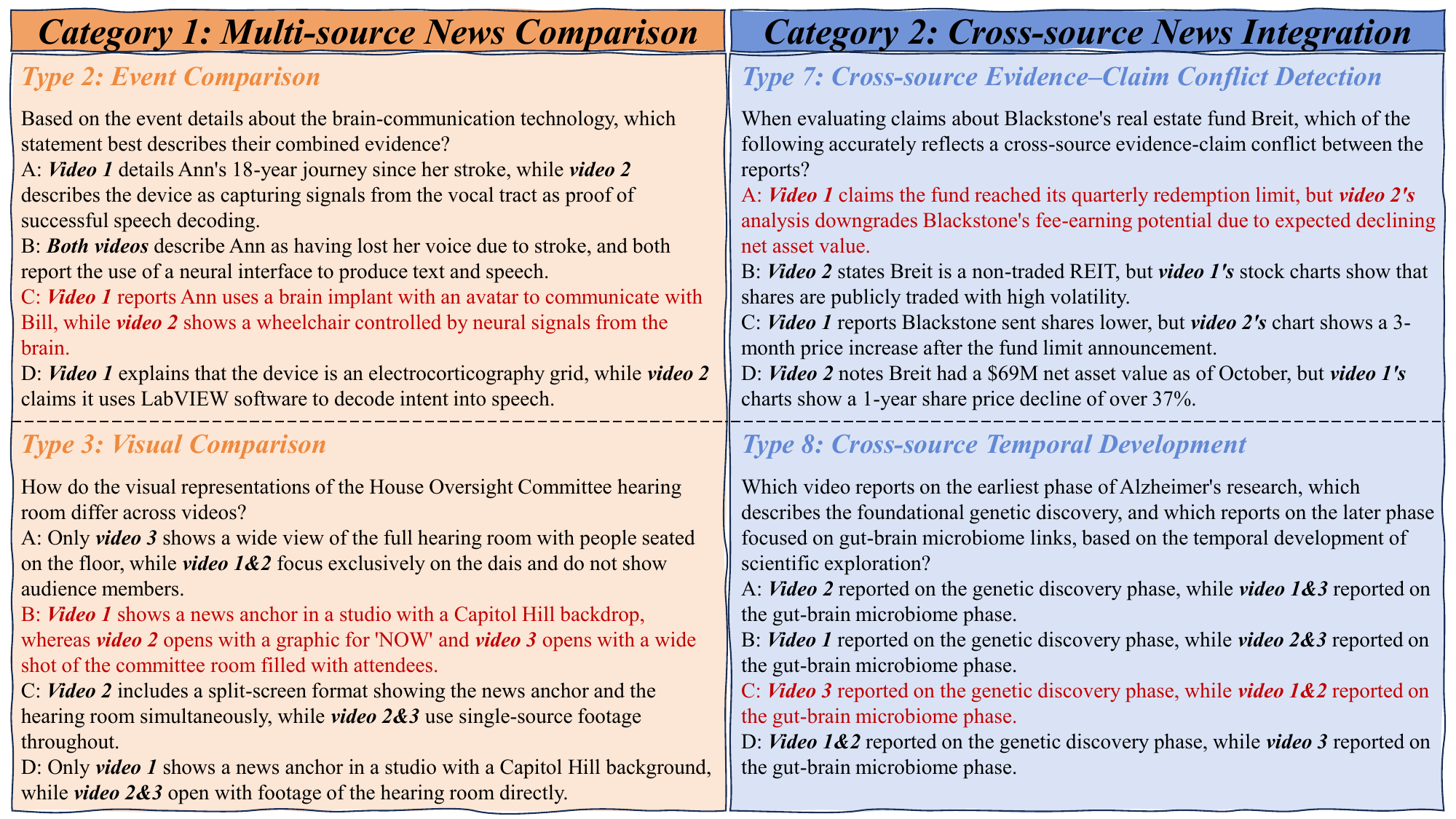}
    \caption{Illustrative examples of generated question types.
The left side shows representative questions of Category 1 (Multi-source News Comparison), which focus on comparing evidence and visuals across videos. The right side shows Category 2 (Cross-source News Integration), requiring cross-video and cross-modal reasoning to detect conflicts and trace temporal developments. Correct answers are highlighted in red. Full examples for all types are provided in Appendix \ref{apd:sample_qa}.}
\label{fig:samples}
\end{figure*}

%% file: sections/experiment.tex
\section{Experiment}

\subsection{Dataset Statistics} 

VNU-Bench contains 429 news groups of 1405 videos in total, with an average video length of 3.65 min, where the video length distribution can be found in Figure~\ref{fig:length_distribution} (a), and the significant video sources can be found in Figure~\ref{fig:length_distribution} (b).  
The dataset contains 2,501 QAs, filtered by quality control from 6,374 draft QAs.  Figure~\ref{fig:domain_distribution} shows the distribution of QAs in different news domains. 
\input{tables_figures_tex/length_distribution}
\input{tables_figures_tex/domain_distribution}

\input{tables_figures_tex/api_results}
\input{tables_figures_tex/domain_result}

\subsection{Evaluation}
We evaluated both closed-source and open-source multimodal large language models (MLLMs) on VNU-Bench for their capability of multi-sourced, multi-modal news understanding. The closed-source models include Gemini-2.5-Pro, Claude-4.5-sonnet and Gemini-2.5-flash. Note that we do not include the models (e.g., GPT-5) used in QA generation.  
For open-source evaluation, we select 12 representative MLLMs, including Qwen2.5-VL-7B~\cite{bai2025qwen2}, InternVL3-8B~\cite{zhu2025internvl3exploringadvancedtraining}, InternVL3.5-8B~\cite{wang2025internvl3_5}, LLaVA-NeXT-7B~\cite{liu2024llavanext}, LLaVA-OneVision-Qwen2-7B~\cite{li2024llava}, the Qwen3 series (Qwen3-VL-8B and Qwen3-VL-30B)~\cite{bai2025qwen3vltechnicalreport}, GLM-4.1V-9B~\cite{vteam2025glm45vglm41vthinkingversatilemultimodal}, and the MiniCPM series (MiniCPM-V-2.6-8B and MiniCPM-V-4.5-9B)~\cite{yu2025minicpmv45cookingefficient}.

We adopted accuracy as the evaluation metric under a zero-shot setting. For each video, we sample 4 top-ranked frames (see Section~\ref{sec:col}). 
All frames were resized such that the longer side was limited to 540 pixels, while the shorter side was scaled proportionally. The transcripts were cleaned  to remove invalid or noisy characters. For those that exceed 1,000 characters, we summarized them to 1,000 characters each. For each video, its transcript and selected frames were concatenated into a unified input block. Additional implementation details are provided in the Appendix \ref{apd:experiment_detail}.

\subsection{Results}
Table~\ref{tab:vnu_closed_open} reports the performance of both closed-source and open-source MLLMs on VNU-Bench, from which several observations can be drawn.

(1) \textbf{Closed-source models outperform open-source models.} Gemini-2.5-Pro achieves the best overall performance with an average accuracy of 60.17\% over the 10 question types, while Claude-4.5-sonnet ranks second, trailing by more than 2\%. Among open-source models, Qwen3-VL-30B performs the best. Generally speaking, a significant gap on average accuracy remains between closed-source models and open-source counterparts, except for the case of Gemini-2.5-flash underperforming when compared to Qwen3-VL-30B and MiniCPM-V-4.5-9B.

(2) \textbf{VNU-Bench is highly challenging.} Even the strongest closed-source model achieves only 60.17\% average accuracy, indicating a substantial gap toward robust multi-source news understanding. Although Gemini-2.5-Pro demonstrates competitive overall performance, it still struggles on some question types (notably T7), highlighting the intrinsic difficulty of coherent multi-video reasoning across heterogeneous sources.

(3) \textbf{Integration questions are substantially more challenging than comparison questions.}
Although both categories of questions require reasoning over multiple news sources, models generally achieve higher accuracy on T1--T5 than on T6--T10, indicating a clear difficulty gap between multi-source comparison and cross-source integration. T1--T5 focus on comparing how different outlets report the same story, including their claims, event details, visual evidence, narrative angles, and multimodal presentation. These questions emphasize alignment and contrast across sources without requiring the construction of a globally coherent reasoning chain.
In contrast, T6--T10 demands deeper cross-source integration, where models must combine heterogeneous textual and visual evidence across outlets, resolve conflicts, track temporal development, and unify narratives.
As a result, most evaluated models exhibit a pronounced performance drop on integration questions, with T7 being particularly challenging.

(4) \textbf{The scaling law is preserved.} Within the Qwen3-VL family, increasing model size from 8B to 30B improves performance across nearly all question types, with Qwen3-VL-30B being the strongest open-source model under evaluation, achieving 56.14\% average accuracy.

(5) \textbf{Architectural improvements also matter.} Newer model versions consistently outperform earlier ones with comparable parameter counts. For example, InternVL3.5-8B significantly outperforms InternVL3-8B, suggesting that advances in architecture and training strategies contribute beyond model scaling alone.

\subsection{Further Analysis}

(1) \textbf{Domain Analysis.}
Figure~\ref{fig:domain_radar} presents a fine-grained, domain-wise comparison of representative MLLMs across five major news domains:  \emph{World}, \emph{US \& Canada}, \emph{Business \& Economy}, \emph{Science \& Technology}, \emph{Climate \& Environment}, \emph{Health}, and \emph{Culture \& Arts}. 
Overall, Claude-4.5-sonnet (red line) achieves the highest accuracy across most question types and news domains. Among different domains, \emph{Business \& Economy} and \emph{Climate \& Environment} exhibit relatively consistent difficulty across question types, whereas \emph{Health} and \emph{World} show higher variance, indicating greater domain-specific reasoning challenges.

(2) \textbf{Frame and Resolution Analysis.}
\input{tables_figures_tex/frames_resolution_analysis}
To further analyze the effects of the number of frames and input image resolution, we conduct two ablation studies. Figure~\ref{fig:qwen_ablation}(a) shows that increasing the number of frames initially improves accuracy; however, performance peaks when six frames are used, and then saturates or slightly degrades when the number of frames is increased further. This behavior may be attributed to the adverse effects of excessively long context lengths in MLLMs~\cite{liu2024lost}. Figure~\ref{fig:qwen_ablation}(b) demonstrates that increasing input image resolution consistently improves accuracy, suggesting that higher visual fidelity provides more informative visual cues for reasoning.



%% file: tables_figures_tex/length_distribution.tex

\begin{figure}[t]
    \centering
    \begin{subfigure}[b]{0.48\linewidth}
        \centering
        \includegraphics[width=\linewidth]{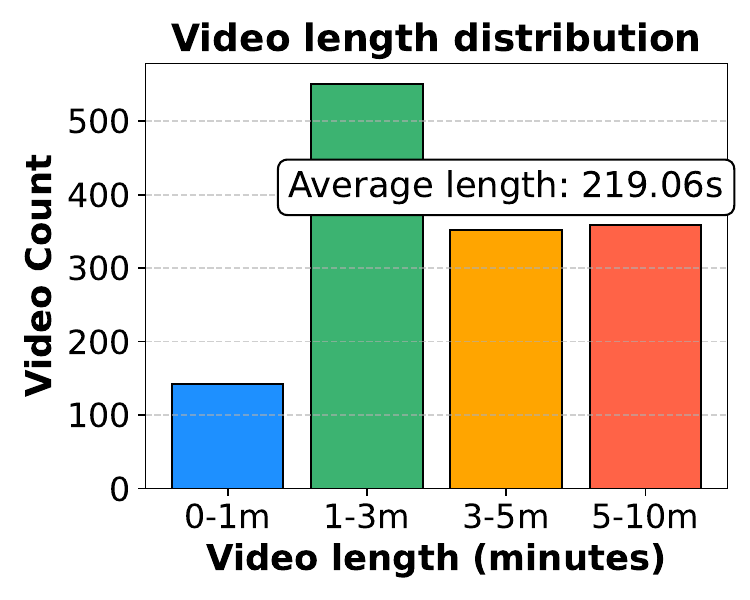}
        \caption{}
        \label{fig:qwen_frames}
    \end{subfigure}
    \hfill
    \begin{subfigure}[b]{0.48\linewidth}
        \centering
        \includegraphics[width=\linewidth]{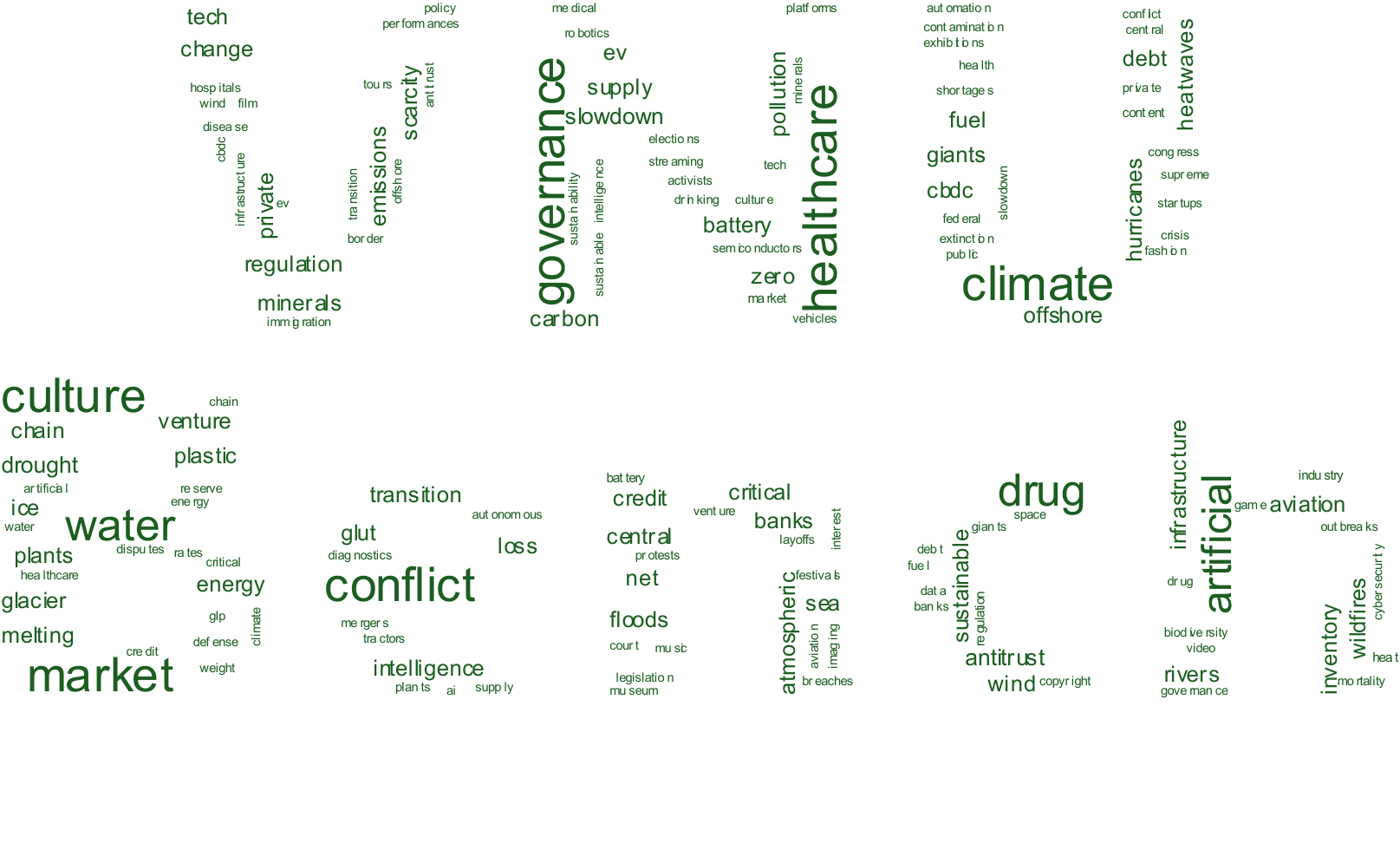}
        \caption{}
        \label{fig:qwen_resolution}
    \end{subfigure}

    \caption{
    (a) Video length distribution and (b) significant organization distribution of VNU-Bench, where the names have to be zoomed in to see.}
    \label{fig:length_distribution}
\end{figure}


%% file: tables_figures_tex/domain_distribution.tex
\begin{figure}[t]  
    \centering
    \includegraphics[width=0.9\linewidth]{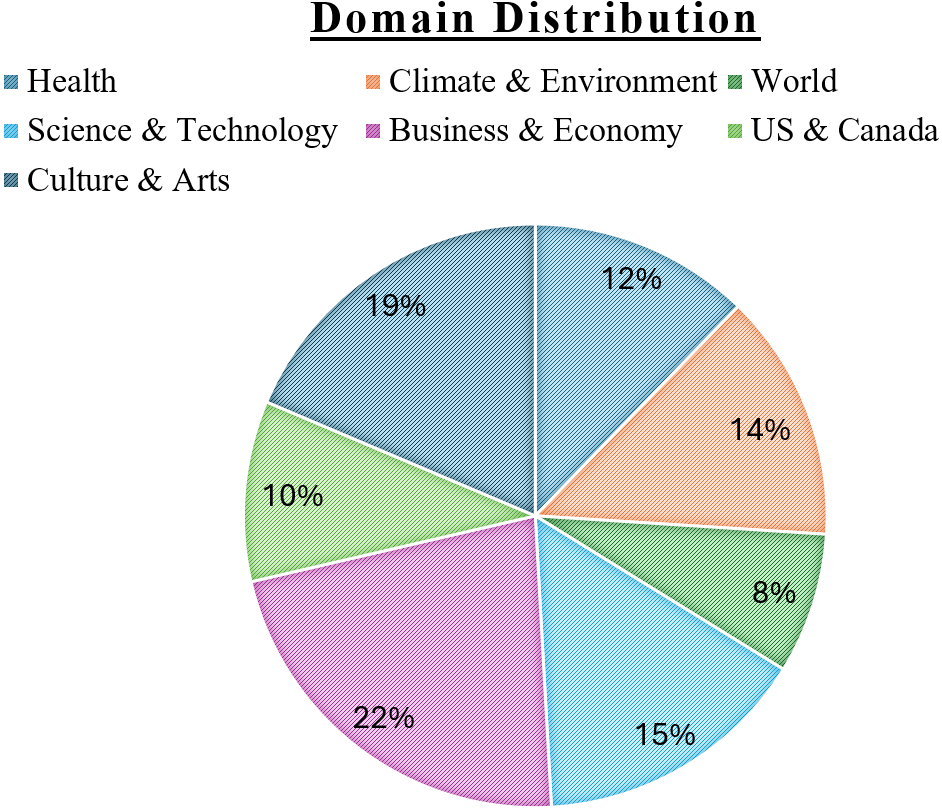}
    \caption{Domain distribution of VNU-Bench.}
    \label{fig:domain_distribution}
\end{figure}

%% file: tables_figures_tex/api_results.tex
\begin{table*}[t]
\centering
\small
\resizebox{\linewidth}{!}{
\begin{tabular}{l|cccccc|cccccc|c}
\hline
\multirow{2}{*}{\textbf{Model}} 
& \multicolumn{6}{c|}{\textbf{Comparison}}
& \multicolumn{6}{c|}{\textbf{Integration}}
& \textbf{Overall} \\
\cline{2-14}

& \textbf{T1} & \textbf{T2} & \textbf{T3} & \textbf{T4} & \textbf{T5} & \textbf{Avg}
& \textbf{T6} & \textbf{T7} & \textbf{T8} & \textbf{T9} & \textbf{T10} & \textbf{Avg}
& \textbf{Avg} \\\hline

\multicolumn{14}{c}{\textbf{Closed-source Models}} \\
\hline

Gemini-2.5-pro
& 61.64 & \textbf{59.53} & 66.67 & 57.69 & \textbf{66.54} & \textbf{62.51}
& 60.02 & 48.26 & \textbf{59.94} & \textbf{56.52} & \textbf{63.86} & \textbf{57.52}
& \textbf{60.17} \\

Claude-4.5-sonnet
& \textbf{63.24} & 52.58 & \textbf{69.00} & 64.80 & 60.57 & 62.00
& 60.08 & \textbf{51.26} & 52.47 & 55.74 & 56.22 & 55.27
& 58.89 \\

Gemini-2.5-flash
& 60.93 & 46.70 & 66.67 & \textbf{66.40} & 61.22 & 60.38
& 45.96 & 42.60 & 53.91 & 46.90 & 48.64 & 47.60
& 54.44 \\

\hline
\multicolumn{14}{c}{\textbf{Open-sourced Models}} \\
\hline

Qwen3-VL-30B
& 55.88 & 51.20 & 67.00 & 59.60 & 60.22 & 58.84
& \textbf{61.47} & 47.48 & 51.12 & 49.18 & 56.22 & 53.01
& 56.14 \\

Qwen3-VL-8B
& 56.99 & 47.42 & 62.67 & 57.60 & 61.65 & 57.26
& 60.63 & 47.48 & 41.70 & 47.13 & 52.36 & 50.33
& 54.06 \\

Qwen2.5-VL-7B
& 53.68 & 49.14 & 59.00 & 57.20 & 55.20 & 54.81
& 49.22 & 39.08 & 47.53 & 45.08 & 46.78 & 45.57
& 50.54 \\

MiniCPM-V-4.5-9B
& 56.43 & 53.45 & 56.80 & 58.48 & 62.36 & 57.50
& 49.09 & 48.30 & 49.08 & 54.58 & 60.20 & 52.25
& 54.97 \\

MiniCPM-V-2.6-8B
& 46.88 & 46.51 & 48.93 & 43.51 & 52.46 & 47.66
& 48.54 & 35.28 & 46.20 & 43.39 & 50.76 & 44.83
& 46.39 \\

GLM-4.1V-9B
& 52.94 & 46.05 & 56.00 & 54.40 & 57.35 & 53.30
& 46.51 & 48.74 & 47.53 & 48.77 & 49.36 & 48.16
& 50.93 \\

InternVL3.5-8B
& 52.94 & 44.27 & 57.67 & 57.20 & 50.54 & 52.12
& 49.61 & 38.66 & 44.39 & 46.67 & 54.51 & 46.77
& 49.92 \\

InternVL3-8B
& 52.84 & 44.30 & 49.00 & 44.20 & 55.48 & 49.16
& 49.06 & 45.54 & 44.50 & 46.08 & 36.62 & 44.36
& 47.02 \\

LLaVA-Next-7B
& 48.16 & 37.80 & 41.67 & 50.40 & 49.10 & 45.19
& 44.19 & 37.39 & 30.04 & 42.21 & 46.78 & 40.30
& 42.93 \\

LLaVA-OneVision
& 35.29 & 35.40 & 38.67 & 38.00 & 42.65 & 38.00
& 43.02 & 34.87 & 39.46 & 34.43 & 35.62 & 37.54
& 37.79 \\

\hline
\end{tabular}
}
\caption{Performance comparison on VNU-Bench across Tasks T1--T10.
Comparison covers T1--T5 and Integration covers T6--T10.
Avg denotes overall accuracy (\%).}
\label{tab:vnu_closed_open}
\end{table*}

%% file: tables_figures_tex/domain_result.tex
\begin{figure*}[t]  
    \centering
    \includegraphics[width=0.9\linewidth]{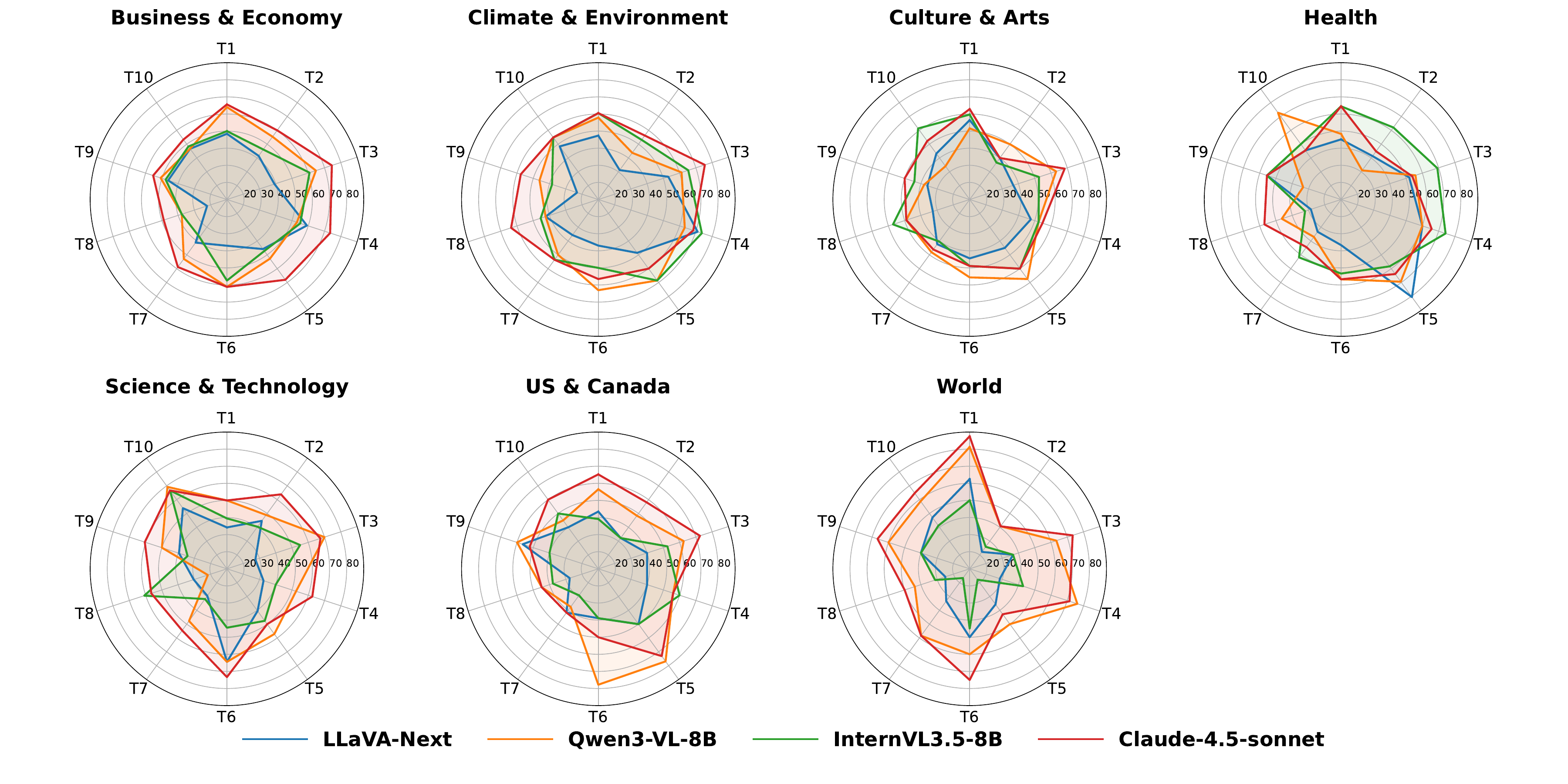}
    \caption{Domain analysis: model accuracy on news of different domains.}
    \label{fig:domain_radar}
\end{figure*}

%% file: tables_figures_tex/frames_resolution_analysis.tex
\begin{figure}[t]
    \centering
    \begin{subfigure}[b]{0.48\linewidth}
        \centering
        \includegraphics[width=\linewidth]{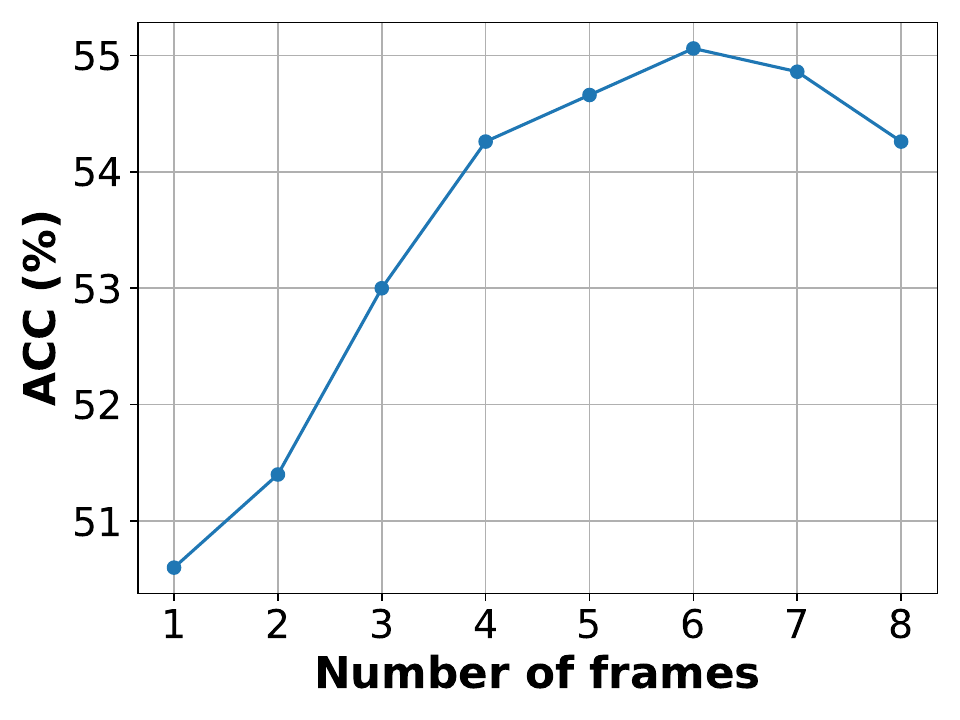}
        \caption{}
        \label{fig:qwen_frames}
    \end{subfigure}
    \hfill
    \begin{subfigure}[b]{0.48\linewidth}
        \centering
        \includegraphics[width=\linewidth]{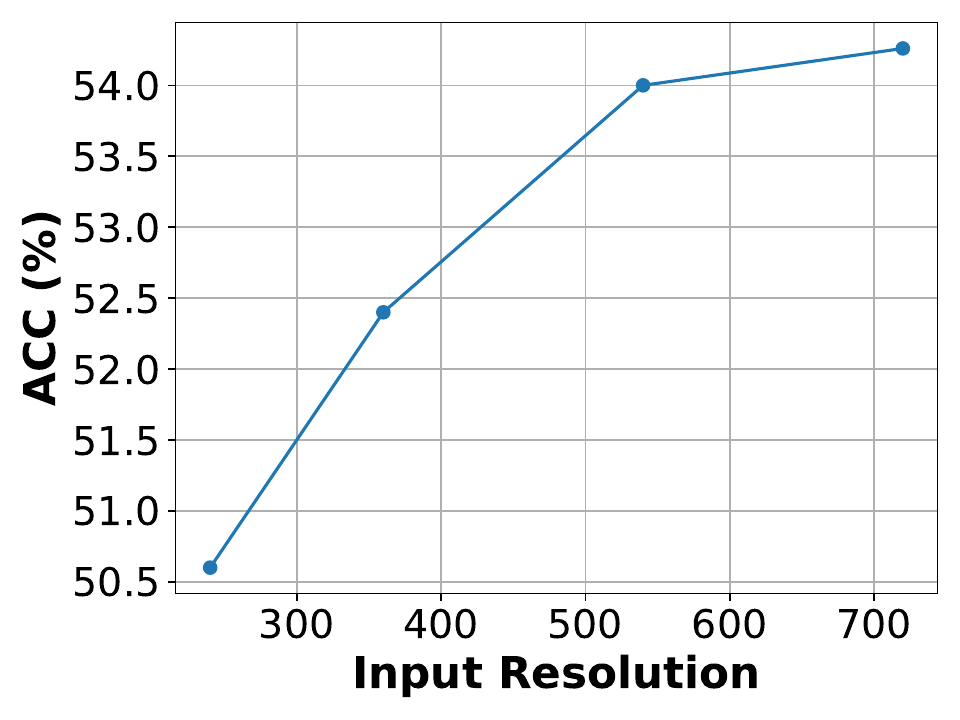}
        \caption{}
        \label{fig:qwen_resolution}
    \end{subfigure}

    \caption{Frame and resolution analysis on Qwen3-VL-8B:
     (a) accuracy with respect to number of frames, and (b) accuracy with respect to input image resolution.}
    \label{fig:qwen_ablation}
\end{figure}

%% file: sections/related_works.tex
\section{Related Works}

\textbf{Specific-Task News Datasets.}
Specific-task news datasets are designed to evaluate isolated capabilities in news understanding by focusing on one specific objective.
Early efforts in the news domain, such as \textit{VisualNews}~\citep{liu2021visual} and \textit{NewsClipping}~\citep{luo2021newsclippings}, which study image--text grounding, captioning, and summarization in news imagery.
Subsequent datasets target more specialized news-oriented problems:
\textit{MOCHEG}~\citep{yao2023end} and \textit{AVERIMATEC}~\citep{cao2025averimatec} focus on multimodal fact checking and verification;
\textit{NewsNet}~\citep{wu2023newsnet} addresses news video segmentation;
and \textit{FakeSV}~\citep{qi2023fakesv} and \textit{EvidSV}~\citep{zeng2025understand} study fake news detection and claim verification.

\textbf{General-purpose News Video Datasets}. Current benchmarks for general-purpose news video understanding primarily evaluate single-source, intra-video comprehension, where multimodal reasoning is limited to aligning narration, visuals, and on-screen text within a single report.
VideoQA~\citep{videoqa} first extends textual QA to the audiovisual domain but focused on local perception without higher-level contextual or temporal abstraction.
Subsequent datasets strengthen multimodal and factual grounding: NewsKVQA~\citep{gupta2022newskvqa} incorporated external knowledge graphs for entity-centric reasoning; NewsVideoQA~\citep{jahagirdar2023watching} emphasizes OCR–ASR–visual integration for text–scene grounding; and ReutersViLNews~\citep{chou2024multi} provides professionally curated captions and retrieval annotations, targeting semantic alignment and narrative generation.

\textbf{General Video Understanding Benchmarks.}
Beyond the news domain, large-scale multimodal benchmarks have extensively studied video understanding.
Early datasets such as TVQA~\citep{lei2018tvqa} and HowTo100M~\citep{miech2019howto100m} focuses on localized perception and clip-level grounding.
More recent multi-task benchmarks aim to holistically evaluate Video-LLMs: Video-Bench~\citep{liu2023videobench} unified QA, captioning, summarization, retrieval, and temporal reasoning; MVBench~\citep{li2023mvbench} expands this paradigm with over twenty subtasks covering perception, reasoning, dialogue, and emotion understanding; and Video-MME~\citep{wang2024videomme} proposes a structured evaluation across perception, cognition, and reasoning.


\textbf{Cross-video Understanding.}
Recently, cross-video reasoning has emerged.
CVBench~\citep{zhu2025cvbench} studies cross-video object and event association in open-domain collections, while CrossVideoQA~\citep{meng2025videoforest} targets person-centric reasoning by aggregating contextual cues across multiple clips.


%% file: sections/conclusion.tex
\section{Conclusion}
We introduce VNU-Bench, the first benchmark for multi-source multi-modal video news understanding, in particular, testing model capability of comparison and integration across reports from multiple outlets covering the same event. Through a hybrid human-model process of question drafting, multi-agent filtering, and human verification, VNU-Bench offers high-quality QAs that span diverse news reasoning types. Our experiments provide the initial insight into the performance of existing MLLMs in their ability of understanding multi-source multi-modal news and show their limitations in some of the more complex questions of cross-source integration.

%% file: sections/limitation.tex
\section{Limitations}

While VNU-Bench is designed to evaluate multi-source and cross-video news understanding, it has several limitations.

First, the dataset focuses on curated news videos collected from a limited set of mainstream news outlets and platforms. As a result, it does not fully capture the diversity of news production styles found in local or non-English media. Models evaluated on VNU-Bench may therefore exhibit different behavior when applied to broader or more heterogeneous news ecosystems.

Second, although we carefully design question types to enforce cross-source dependency, the benchmark is constructed in a multiple-choice format. This formulation simplifies evaluation but may not fully reflect open-ended reasoning, evidence grounding, or narrative generation abilities required in real-world news analysis.

Third, parts of the dataset construction and quality control pipeline rely on large language models for question drafting, filtering, and difficulty estimation. While multiple models and human verification are employed to mitigate bias, residual model-induced preferences or stylistic patterns may still influence the final question distribution.

%% file: sections/appendix.tex



\section{More details about VNU-Bench}
More statistics about our dataset is shown in the Table  \ref{tab:data_distribution}, where we report the number of news groups, total videos, the counts of questions for each task type (T1–T10), and the total number of questions for each news domain.

\input{tables_figures_tex/data_statistics}

\section{Sample Questions of the 10 Types}\label{apd:sample_qa}
This section contains the detailed samples for our 10 types of questions. We provide the video link for questions. We also add screenshots of videos for better understanding. 

\subsection{Layer 1: Multi-source News Comparison}

\begin{itemize}
\item{Type 1: Main claim comparison.} This type of questions test a model's understanding of the main claims in different news reportings about a common story, as well as similarity and contrast in the claims.  

\textbf{Sample Question:}
When comparing the main conclusions of Ran Chen-Global MedTech Expert, Everyday Bioethics Expert, and Versa AI Hub about the same topic of AI in healthcare, which statement best captures how their overarching claims differ? (Screenshots are shown in Figure\ref{fig:type3}.)

\begin{itemize}[leftmargin=1.2em, itemsep=2pt, topsep=2pt, parsep=0pt]

  \item Ran Chen-Global MedTech Expert: \url{https://www.youtube.com/watch?v=fhg6I14Waw0}

  \item Everyday Bioethics Expert:  \url{https://www.youtube.com/watch?v=nUdP-67N2o0}

  \item Versa AI Hub: \url{https://www.youtube.com/watch?v=-cmnIRBZGtw}
  
  \item Techugo Private Limited: \url{https://www.youtube.com/watch?v=Dm-GR1dCzco}

\end{itemize}

\textbf{Multiple Choices:}
A. Both Ran Chen and Versa AI Hub frame AI as requiring regulatory oversight, while Everyday Bioethics Expert focuses only on patient data security and does not address classification or ethical events.\\
B. Everyday Bioethics Expert and Versa AI Hub both warn of data breaches, while Ran Chen claims Mexico’s new rules have no impact on AI privacy or ethical oversight. \\
C. Ran Chen argues AI software is now classified as a medical device in Mexico with risk tiers, while the Everyday Bioethics Expert emphasizes that AI diagnosis demands strong privacy safeguards, and Versa AI Hub states the UAE is hosting forums to align AI with ethical standards. \\
D. All three sources agree AI improves diagnostics, but only Ran Chen and Everyday Bioethics Expert mention regulatory frameworks, while Versa AI Hub dismisses the need for ethics forums.

\textbf{Correct choice:} C

\textbf{Explanation:}
Ran Chen: software now classified as a medical device in Mexico under risk tiers; Everyday Bioethics Expert: AI diagnosis requires strong privacy safeguards and regulations like HIPAA/GDPR; Versa AI Hub: 'The event aims to align AI advancements with healthcare ethical standards'

\hrulefill

\item{Type 2: Event comparison.} This type of questions test a model's understanding of the details of events reported in different news videos about a common story, as well as coherence in these evidences and whether they contradict one another.

\textbf{Sample Question:}
Based on the details of events from NBC News, euronews, and DW News about the same Iran-US standoff, which statement correctly compares their claims about the US military presence and Iran’s stated motivations? (Screenshots are shown in Figure\ref{fig:iran_us_three_sources}.)

\begin{figure}[H]
  \centering

  \begin{subfigure}[t]{0.28\textwidth}
    \centering
    \includegraphics[width=\linewidth]{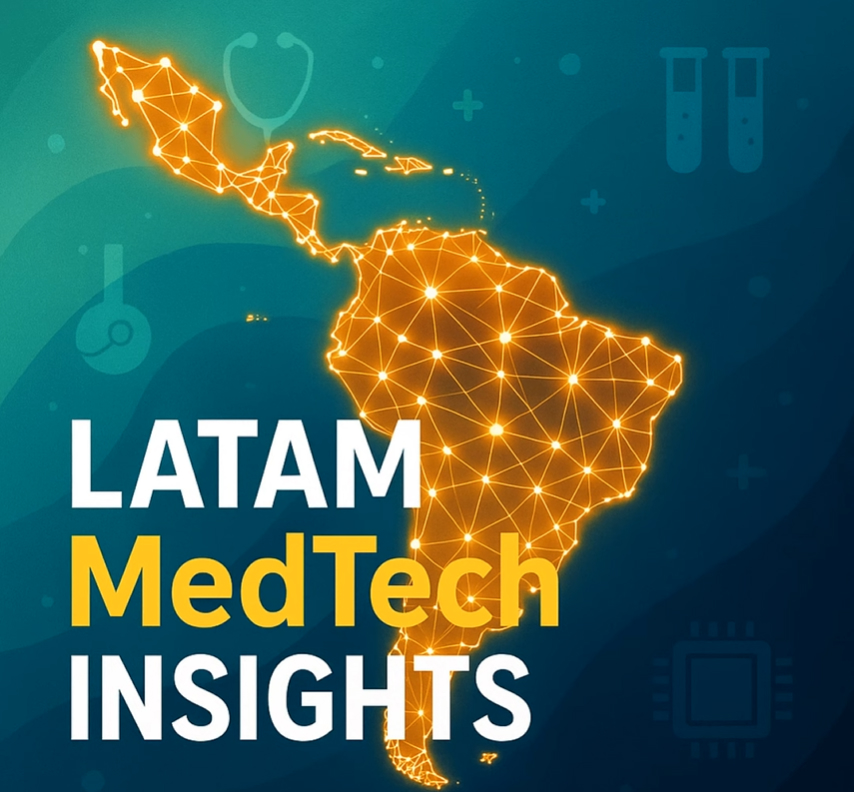}
    \caption{Ran Chen-Global MedTech Expert}
    \label{fig:iran_us_nbc}
  \end{subfigure}
  \hfill
  \begin{subfigure}[t]{0.28\textwidth}
    \centering
    \includegraphics[width=\linewidth]{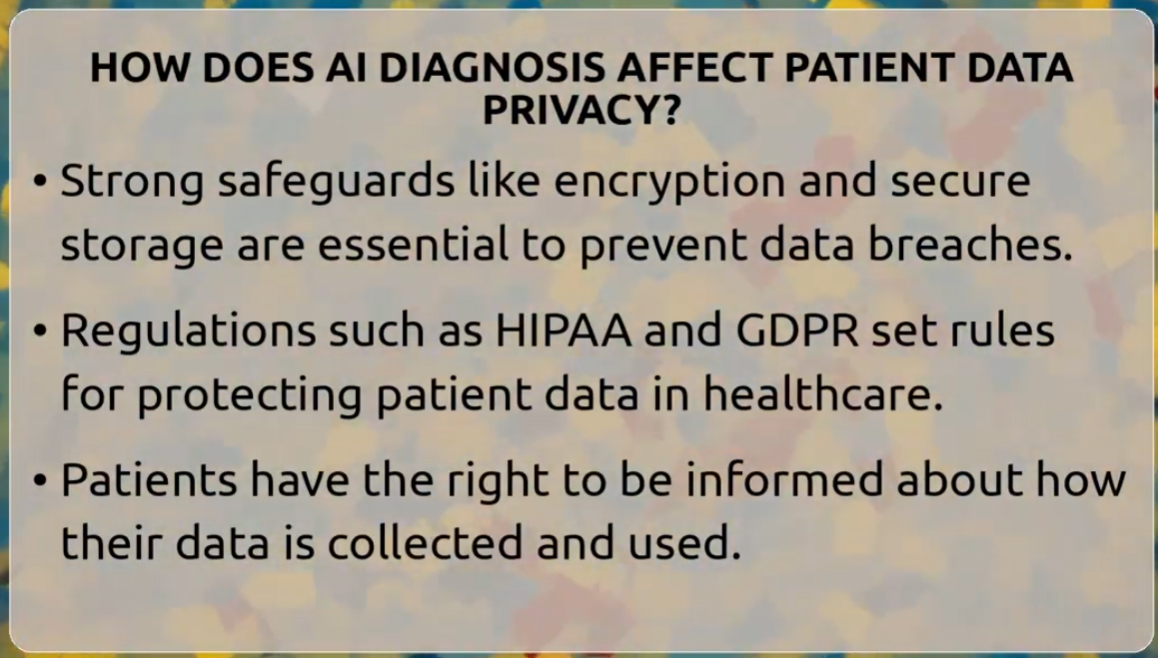}
    \caption{Everyday Bioethics Expert}
    \label{fig:iran_us_euronews}
  \end{subfigure}
  \hfill
  \begin{subfigure}[t]{0.28\textwidth}
    \centering
    \includegraphics[width=\linewidth]{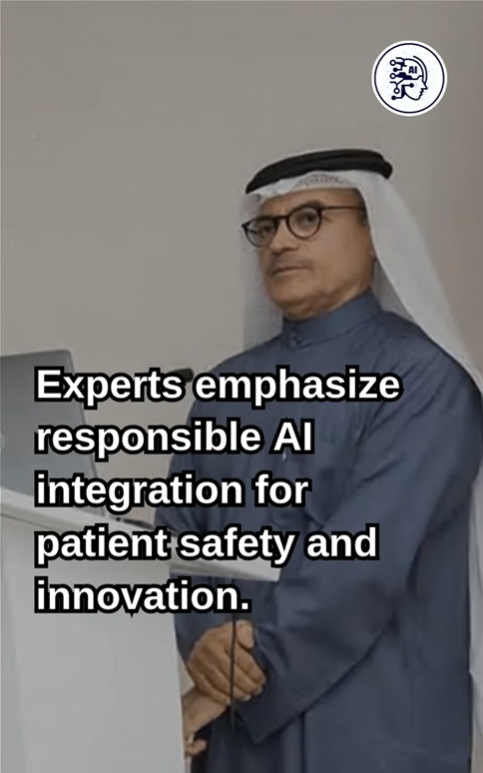}
    \caption{Versa AI Hub}
    \label{fig:iran_us_dw}
  \end{subfigure}
    \hfill
  \begin{subfigure}[t]{0.28\textwidth}
    \centering
    \includegraphics[width=\linewidth]{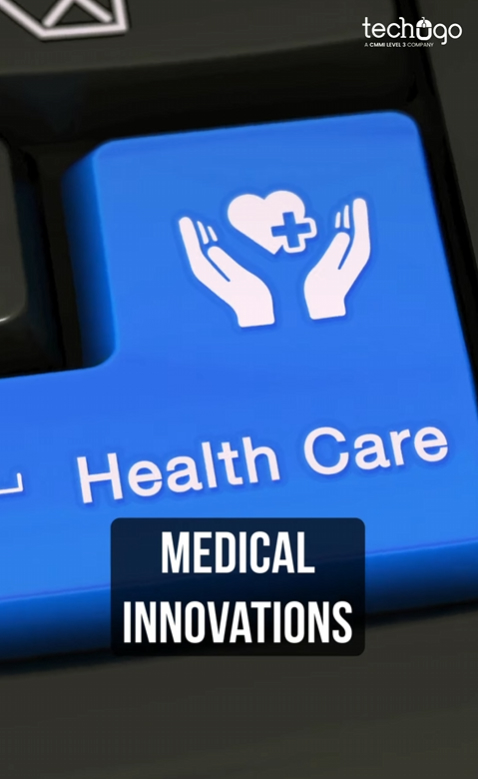}
    \caption{Techugo Private Limited}
    \label{fig:iran_us_dw}
  \end{subfigure}

  \caption{Screen shots for the videos in Type 1 and 3.}
  \label{fig:type3}
\end{figure}

\begin{itemize}[leftmargin=1.2em, itemsep=2pt, topsep=2pt, parsep=0pt]

  \item NBC News: \url{https://www.youtube.com/watch?v=diJ6Et439ao}

  \item euronews:  \url{https://www.youtube.com/watch?v=YecB-ARBnC4}

  \item DW news: \url{https://www.youtube.com/watch?v=LDyIaurlRZs}

\end{itemize}

\textbf{Multiple Choices:}
A. NBC News reports an 'extreme alert' in Israel due to a missile launch, euronews states Iran is testing missiles and intends to close the Strait of Hormuz if sanctions continue, and DW News confirms the US is deploying an assault ship and Patriot missile battery to the Persian Gulf. \\

B. NBC News and DW News both report the US is sending vessels to the Gulf without naming the threat source, while euronews claims Iran is testing missiles and suspending nuclear deal clauses to pressure the remaining JCPAO parties.\\

C.NBC News and euronews both claim Iran is preparing a missile attack against Israel, while DW News states the US deployed naval forces in response without mentioning Iran’s nuclear deal suspensions. \\

D. NBC News and DW News report Iran threatening to close the Strait of Hormuz, while euronews states the US is deploying a naval strike group without specifying Iran’s motivations for missile tests.

\textbf{Correct choice:} A

\textbf{Explanation:}
NBC News reports Iran launching retaliatory missiles and Israel sending an 'extreme alert'; euronews details Iran’s missile tests and its threat to close the Strait of Hormuz if sanctions affect oil exports; DW News specifies the US is deploying the USS Arlington and Patriot system to the Persian Gulf.

\hrulefill
\begin{figure}[H]
  \centering

  \begin{subfigure}[t]{0.32\textwidth}
    \centering
    \includegraphics[width=\linewidth]{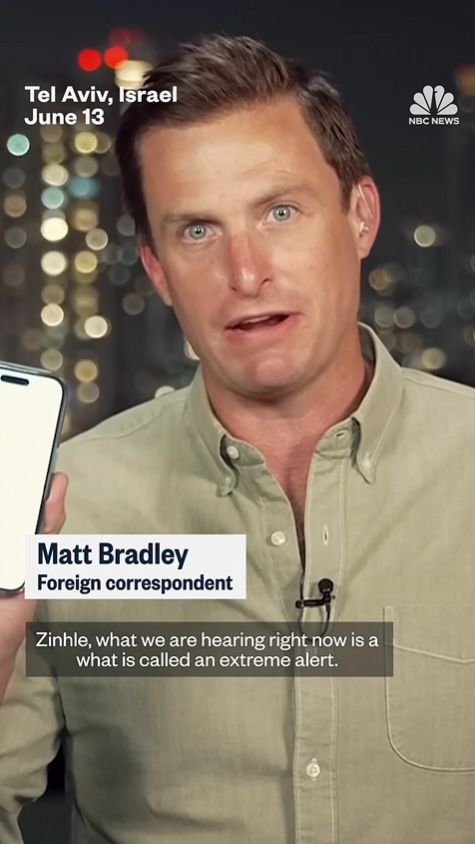}
    \caption{NBC News}
    \label{fig:iran_us_nbc}
  \end{subfigure}
  \hfill
  \begin{subfigure}[t]{0.32\textwidth}
    \centering
    \includegraphics[width=\linewidth]{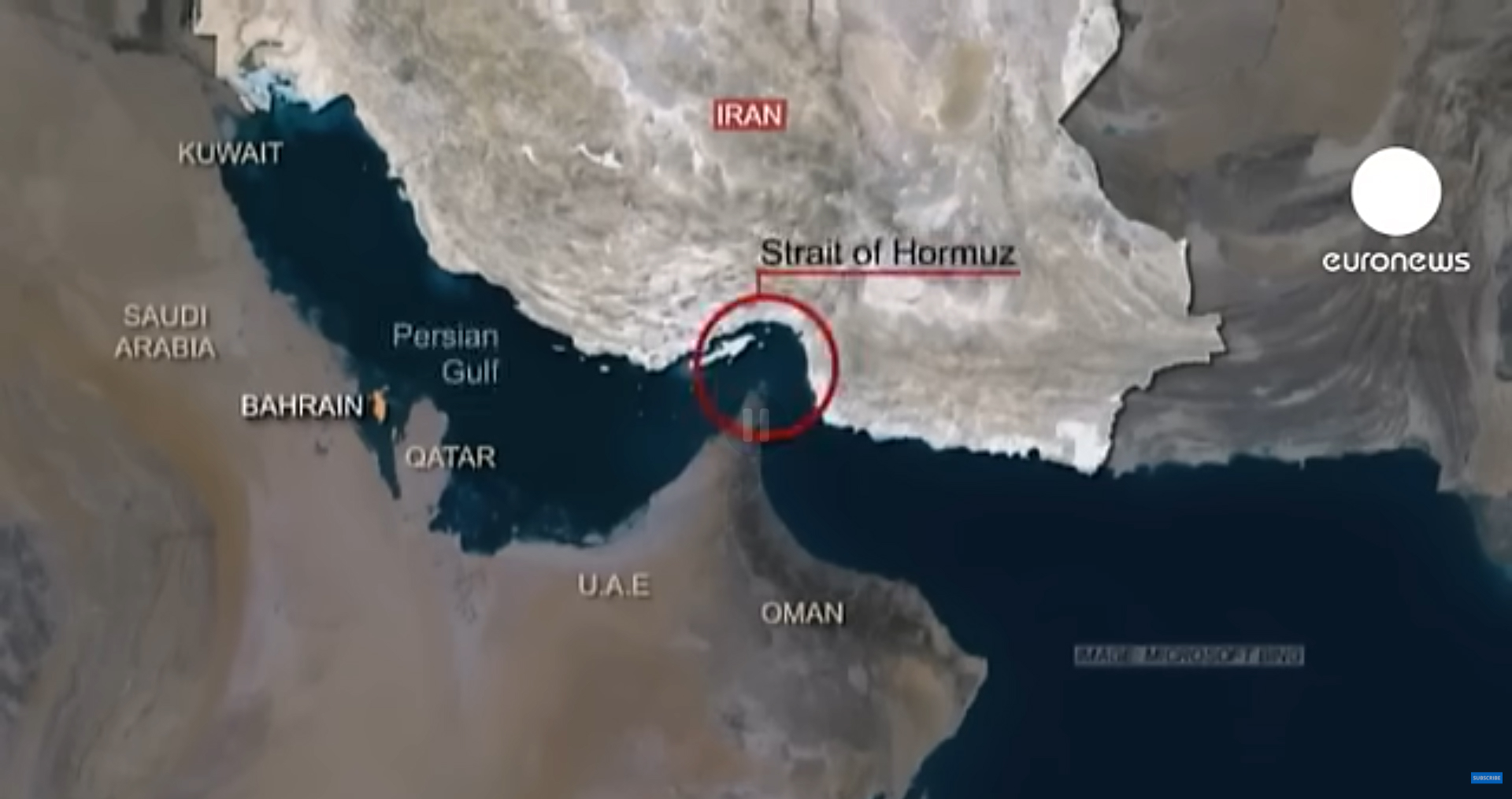}
    \caption{euronews}
    \label{fig:iran_us_euronews}
  \end{subfigure}
  \hfill
  \begin{subfigure}[t]{0.32\textwidth}
    \centering
    \includegraphics[width=\linewidth]{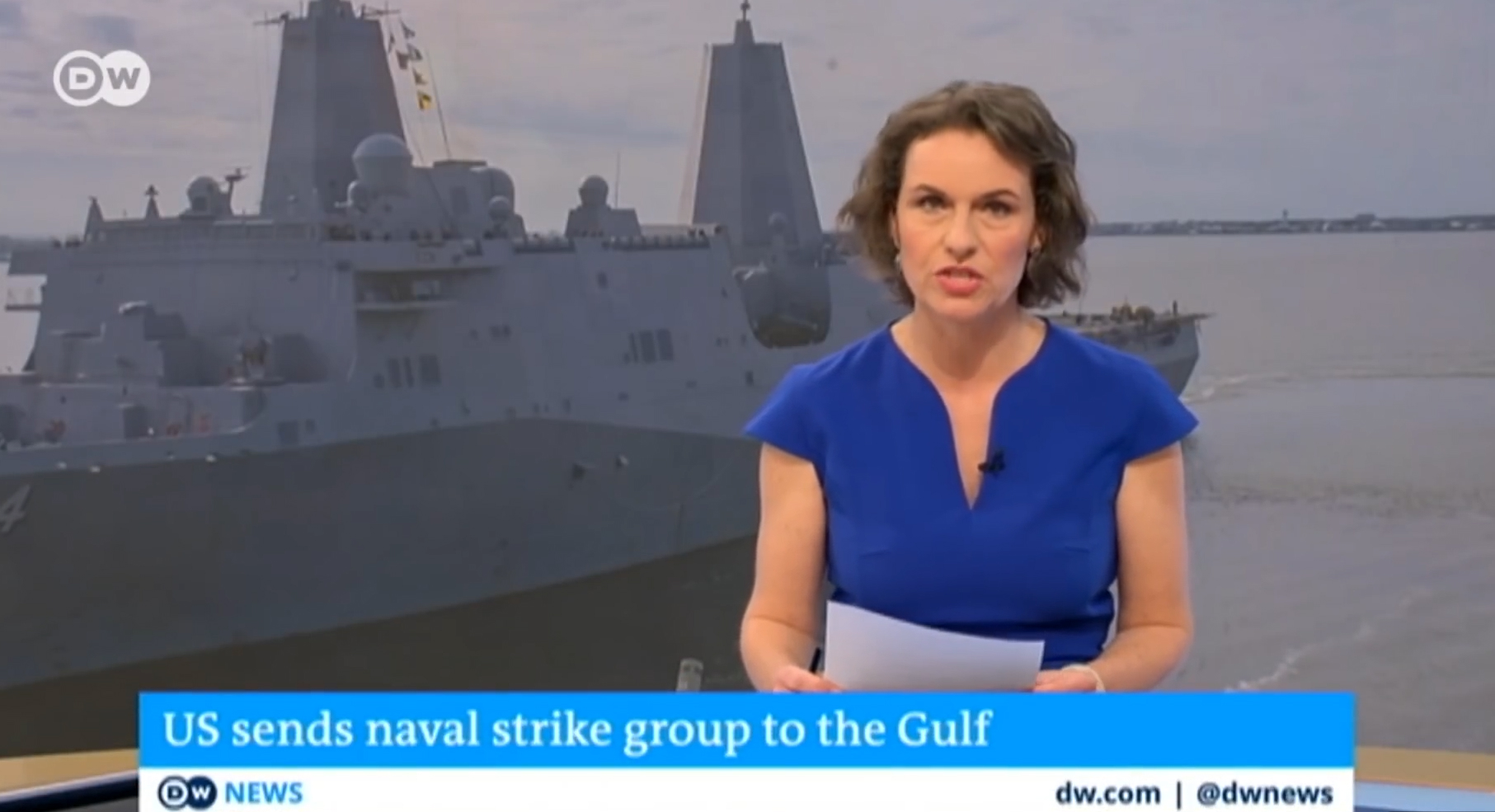}
    \caption{DW News}
    \label{fig:iran_us_dw}
  \end{subfigure}

  \caption{Screen shots for the videos in Type 2.}
  \label{fig:iran_us_three_sources}
\end{figure}

\item{Type 3: Visual comparison.} This type of questions test a model's understanding of information from the images in different news videos about a common story and compare how such information supports the claims in the reportings.   

\textbf{Sample Question:}
Which outlets visually show on-screen text identifying specific AI medical regulations or privacy safeguards, and which do not? (Screenshots are shown in Figure\ref{fig:type3}.)

\begin{itemize}[leftmargin=1.2em, itemsep=2pt, topsep=2pt, parsep=0pt]

  \item Ran Chen-Global MedTech Expert: \url{https://www.youtube.com/watch?v=fhg6I14Waw0}

  \item Everyday Bioethics Expert:  \url{https://www.youtube.com/watch?v=nUdP-67N2o0}

  \item Versa AI Hub: \url{https://www.youtube.com/watch?v=-cmnIRBZGtw}
  
  \item Techugo Private Limited: \url{https://www.youtube.com/watch?v=Dm-GR1dCzco}

\end{itemize}

\textbf{Multiple Choices:}
A. Ran Chen-Global MedTech Expert shows a map of Latin America with 'LATAM MedTech INSIGHTS' text; Everyday Bioethics Expert displays bullet points on 'HIPAA and GDPR'; Versa AI Hub never shows any regulatory text. \\
B. Everyday Bioethics Expert shows regulatory text on 'HIPAA and GDPR'; Versa AI Hub shows a man at a podium with overlay text 'Experts emphasize responsible AI integration'; Ran Chen-Global MedTech Expert never shows regulatory text. \\
C. Ran Chen-Global MedTech Expert and Versa AI Hub show regulatory text related to AI; Everyday Bioethics Expert displays text about 'federated learning' and 'differential privacy' but no jurisdiction-specific regulations. \\
D. Everyday Bioethics Expert and Versa AI Hub display regulatory text such as 'HIPAA' and 'GDPR'; Ran Chen-Global MedTech Expert only shows the title 'LATAM MedTech INSIGHTS' without any regulation references.

\textbf{Correct choice:} B

\textbf{Explanation:}
Everyday Bioethics Expert's frames show text 'HIPAA and GDPR' under privacy safeguards; Versa AI Hub shows overlay text 'Experts emphasize responsible AI integration'; Ran Chen-Global MedTech Expert's visuals are a map with a title, no regulatory text.

\hrulefill

\item{Type 4: Narrative angle comparison.} This type of questions tests a model’s understanding of the narrative angles used by different news outlets when reporting the same story. A narrative angle refers to the interpretive lens or perspective through which an outlet organizes and presents the event—shaping what is emphasized or downplayed, which causes or actors are foregrounded, what values or norms are invoked, and how the meaning or significance of the story is constructed. These questions focus on comparing how outlets highlight certain details of events, interpretations, or concerns while deemphasizing or reframing others.

\textbf{Sample Question:}
How do the reports by Global News, The Canadian Press, and TVC News Nigeria each frame the diplomatic fallout between Canada and India following the alleged killing of a Sikh separatist? (Screenshots are shown in Figure\ref{fig:type4}.)

\begin{figure}[H]
  \centering

  \begin{subfigure}[t]{0.32\textwidth}
    \centering
    \includegraphics[width=\linewidth]{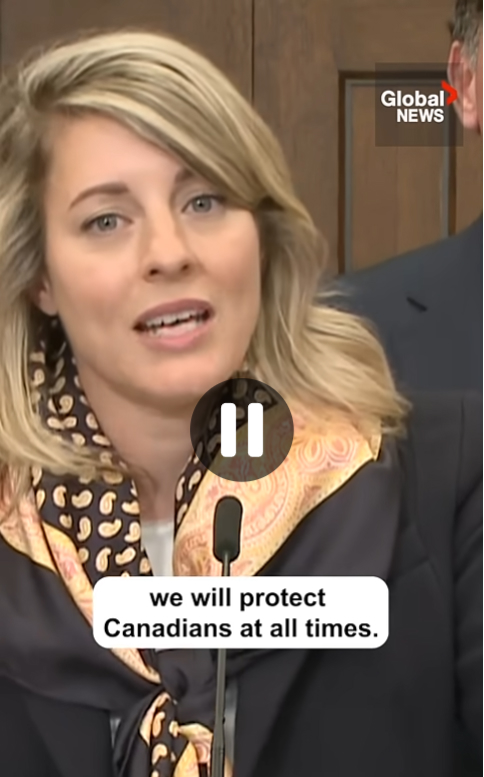}
    \caption{Global News}
    \label{fig:iran_us_nbc}
  \end{subfigure}
  \hfill
  \begin{subfigure}[t]{0.32\textwidth}
    \centering
    \includegraphics[width=\linewidth]{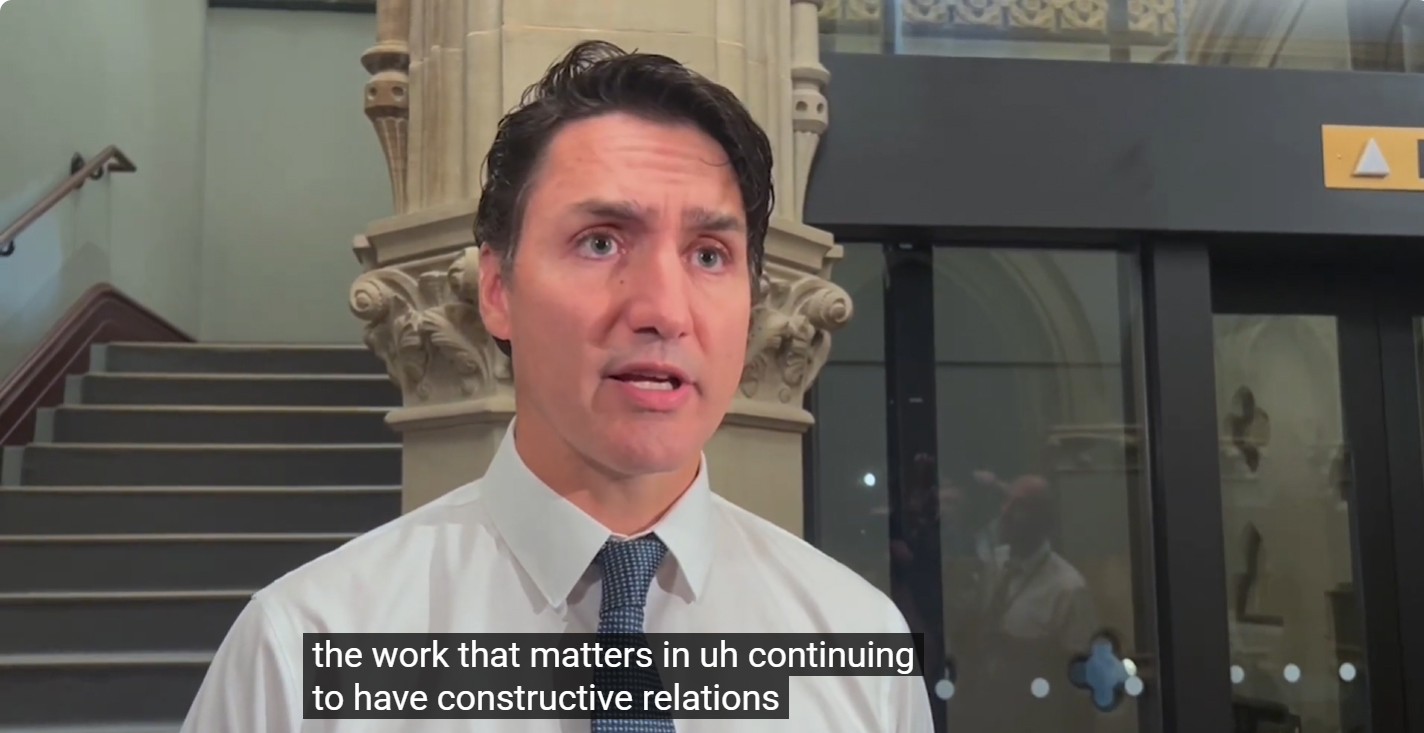}
    \caption{The Canadian Press}
    \label{fig:iran_us_euronews}
  \end{subfigure}
  \hfill
  \begin{subfigure}[t]{0.32\textwidth}
    \centering
    \includegraphics[width=\linewidth]{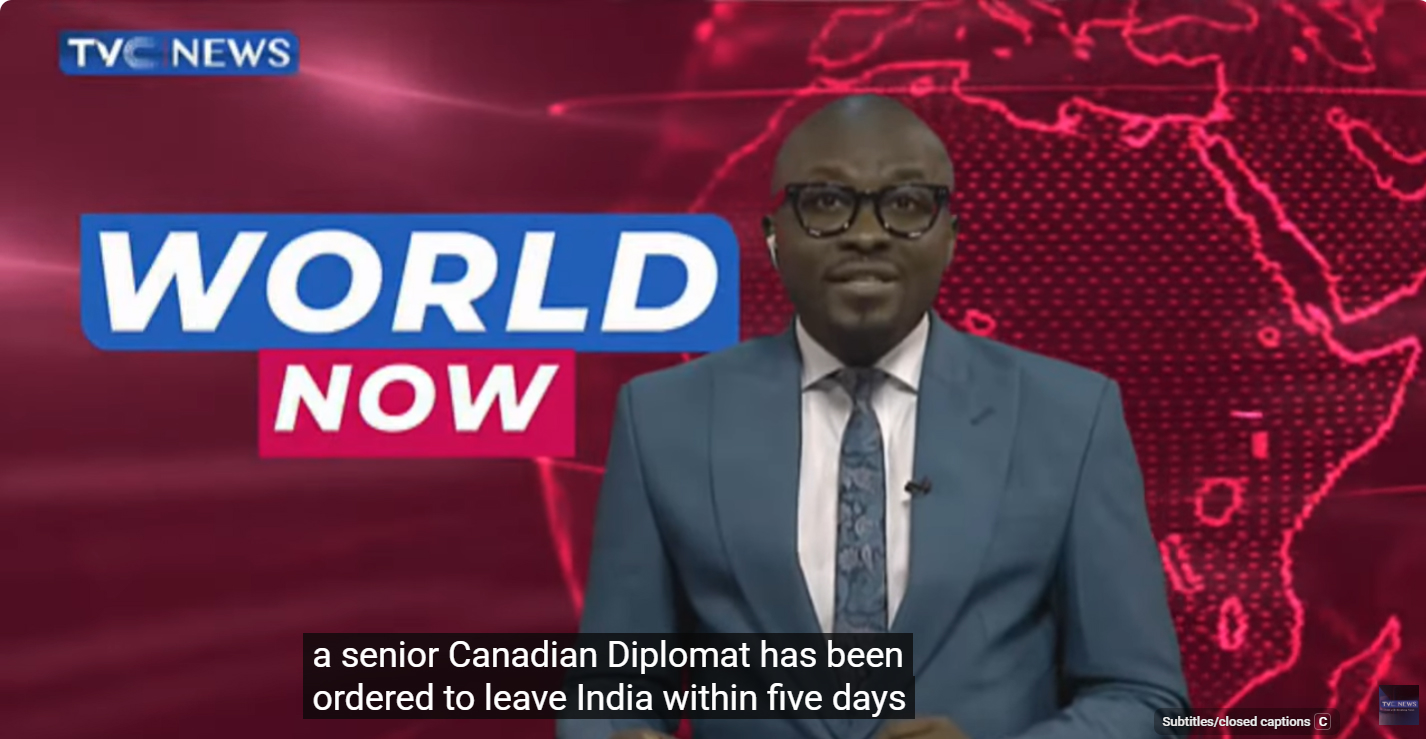}
    \caption{TVC News Nigeria}
    \label{fig:iran_us_dw}
  \end{subfigure}

  \caption{Screen shots for the videos in Type 4.}
  \label{fig:type4}
\end{figure}

\begin{itemize}[leftmargin=1.2em, itemsep=2pt, topsep=2pt, parsep=0pt]

  \item Global News: \url{https://www.youtube.com/watch?v=6k5e5o-F7tU}

  \item The Canadian Press:  \url{https://www.youtube.com/watch?v=eetVv5jZanM}

  \item TVC News Nigeria: \url{https://www.youtube.com/watch?v=yoJ6NWvoY6w}

\end{itemize}

\textbf{Multiple Choices:}
A. Global News frames the expulsion as a necessary assertion of sovereignty guided by clear principles, The Canadian Press presents Trudeau’s stance as cautious and focused on constructive engagement, and TVC News Nigeria emphasizes the bilateral retaliation and escalating geopolitical consequences. \\
B. Global News portrays the situation as a personal crisis for Canada, The Canadian Press views it as an internal diplomatic dispute, and TVC News Nigeria frames it as a symbolic diplomatic incident with minimal long-term impact. \\
C. Global News highlights Canada’s pursuit of truth and public safety as central to its response, The Canadian Press emphasizes the risk of escalation and Canada’s desire to avoid conflict, and TVC News Nigeria focuses on the economic and diplomatic fallout including canceled trade talks. \\
D. Global News frames the events as a routine diplomatic protocol, The Canadian Press treats the expulsion as an isolated incident with no significant consequences, and TVC News Nigeria downplays the tensions and suggests they are overblown by the media.

\textbf{Correct choice:} A

\textbf{Explanation:}
Global News portrays the expulsion as a principle-driven and necessary measure; The Canadian Press characterizes Trudeau’s position as cautious and oriented toward constructive engagement; TVC News Nigeria highlights the reciprocal move and escalating geopolitical repercussions.

\hrulefill

\item{Type 5: Multi-modal presentation comparison.} This type of questions test a model's understanding of multi-modal presentations in different news reportings about a common story and how effectively they integrate audio (textual) and video in presenting their evidences and arguments to support their main claims. 

\textbf{Sample Question:}
How do the different news sources combine narration and visuals to present their coverage of the G7's \$50 billion loan plan for Ukraine? (Screenshots are shown in Figure\ref{fig:type5})

\begin{figure}[H]
  \centering

  \begin{subfigure}[t]{0.32\textwidth}
    \centering
    \includegraphics[width=\linewidth]{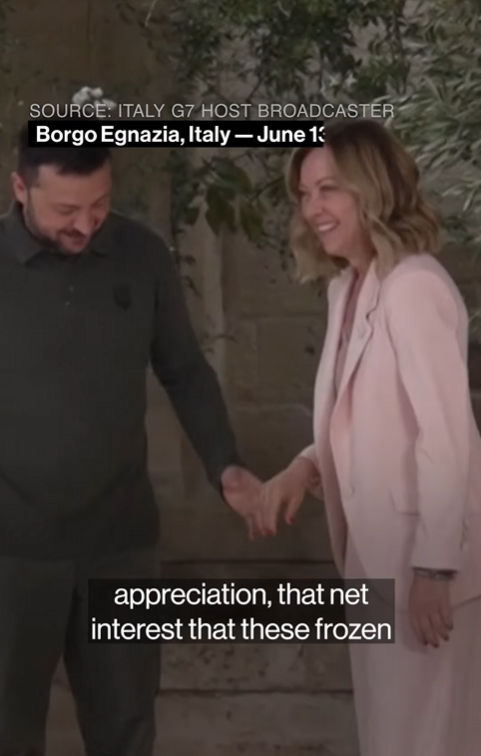}
    \caption{Bloomberg News}
    \label{fig:iran_us_nbc}
  \end{subfigure}
  \hfill
  \begin{subfigure}[t]{0.32\textwidth}
    \centering
    \includegraphics[width=\linewidth]{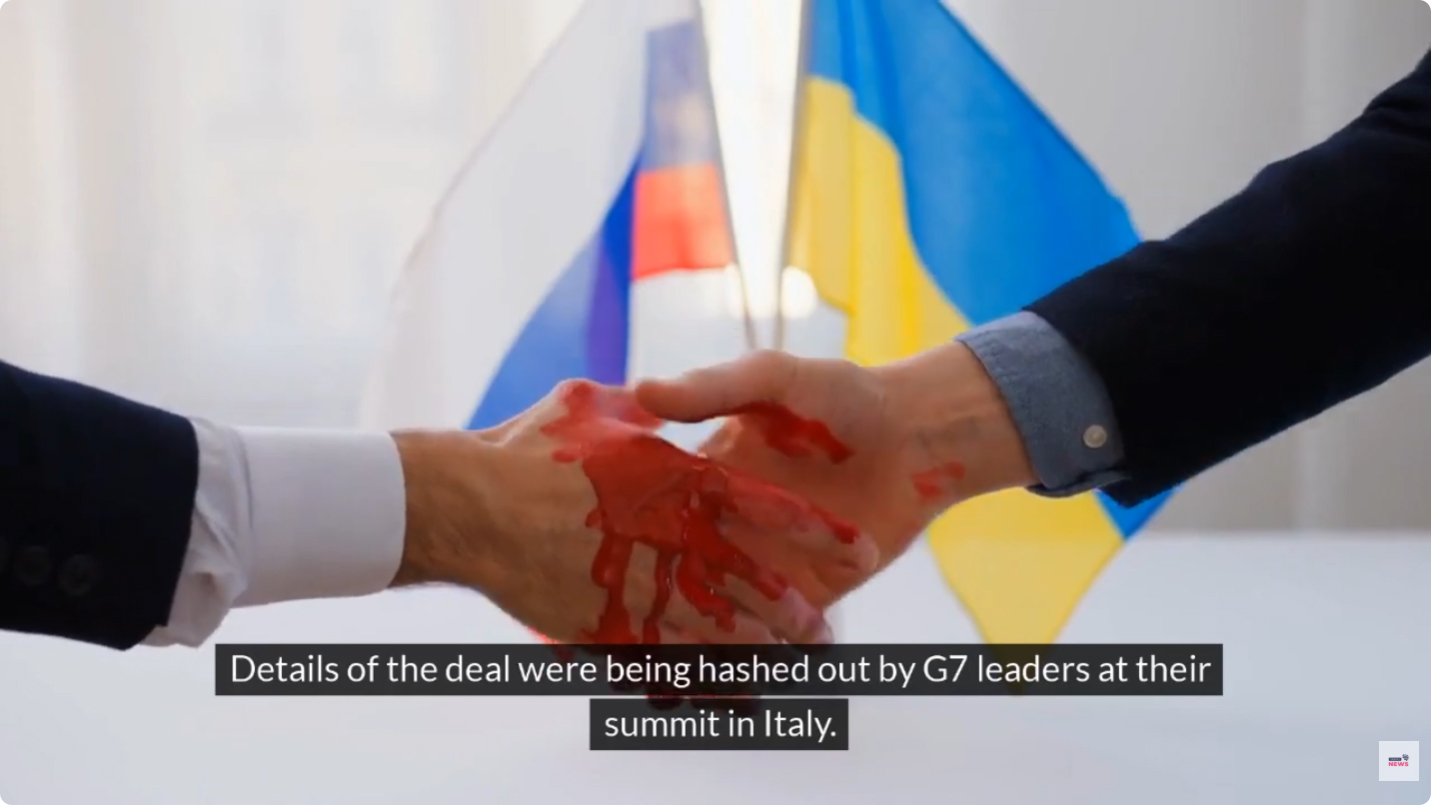}
    \caption{The Economic Times}
    \label{fig:iran_us_euronews}
  \end{subfigure}
  \hfill
  \begin{subfigure}[t]{0.32\textwidth}
    \centering
    \includegraphics[width=\linewidth]{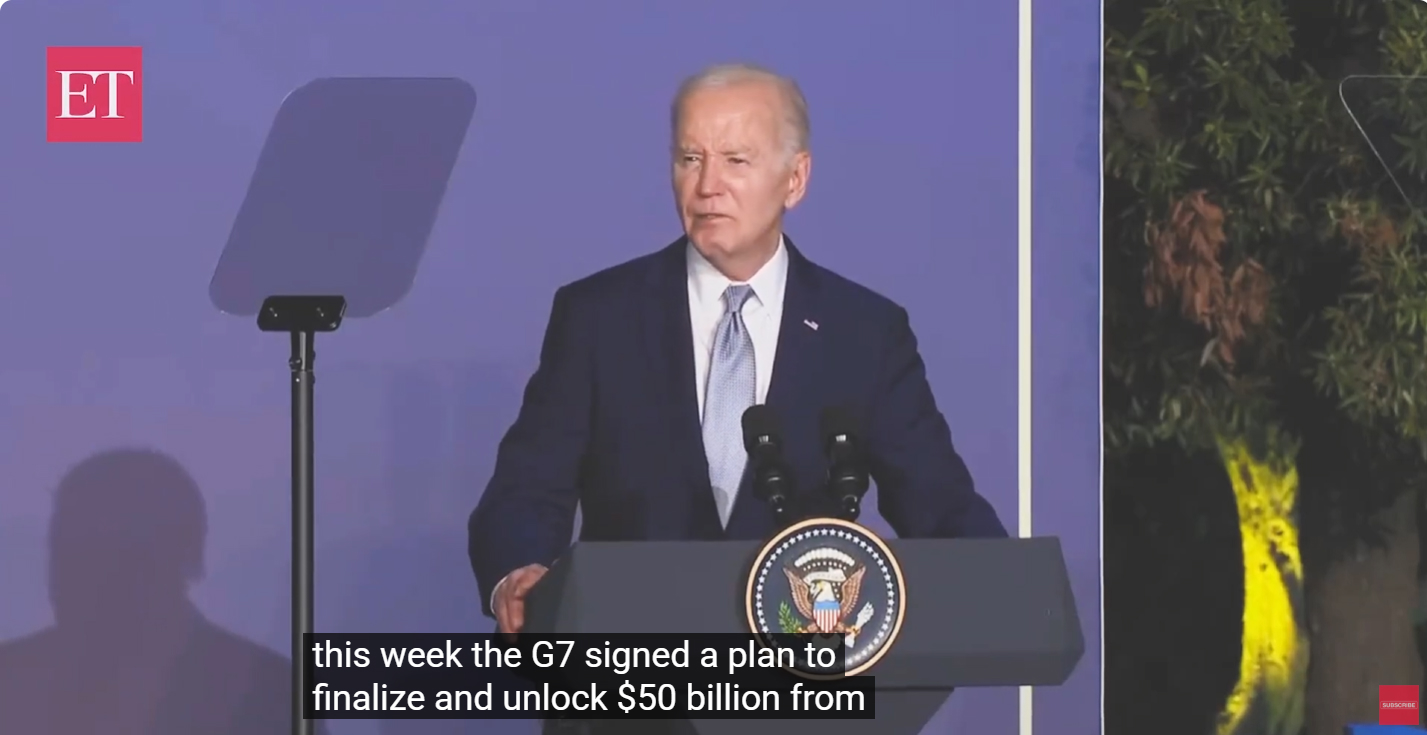}
    \caption{Breaking Now}
    \label{fig:iran_us_dw}
  \end{subfigure}
    \hfill
  \begin{subfigure}[t]{0.32\textwidth}
    \centering
    \includegraphics[width=\linewidth]{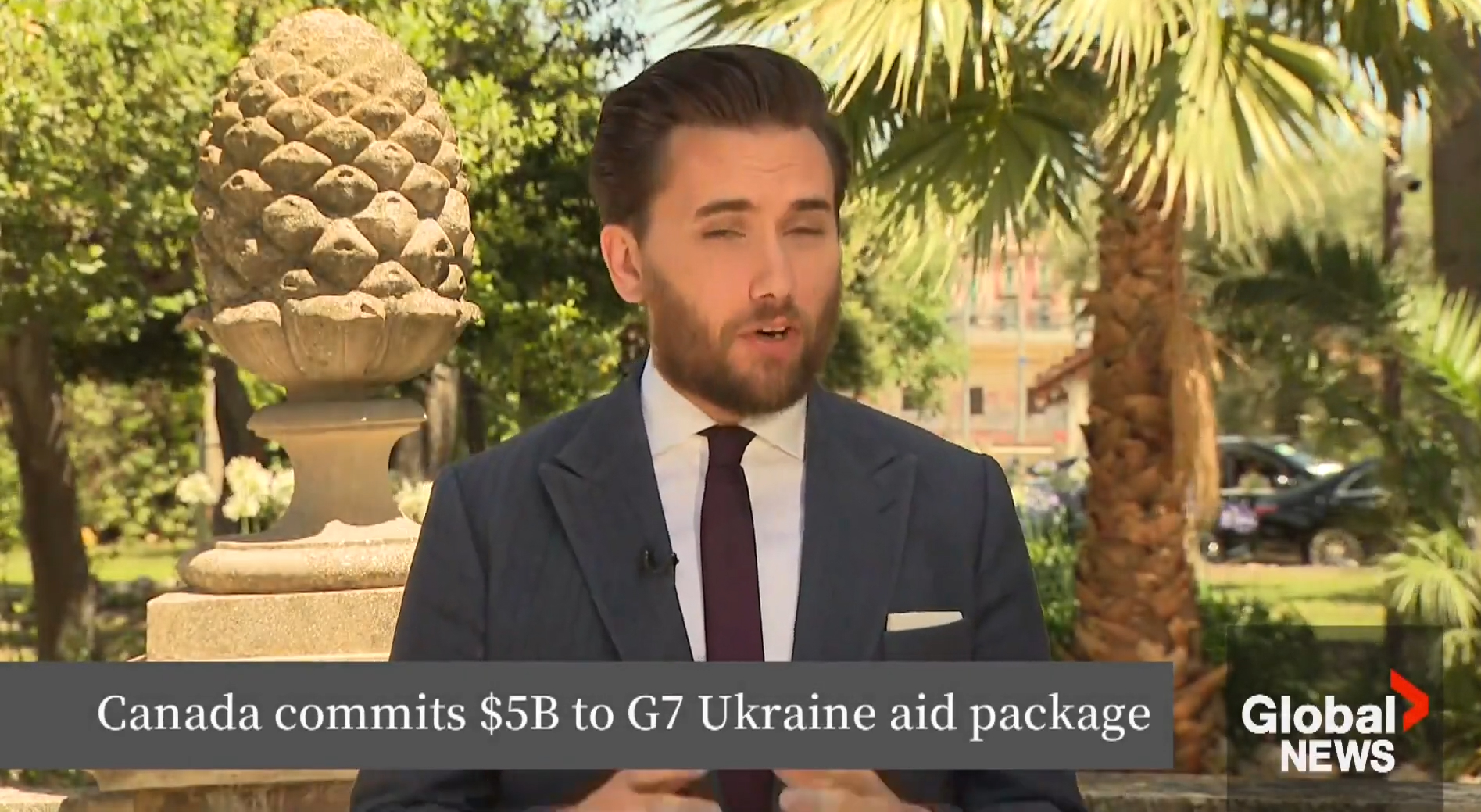}
    \caption{Global News}
    \label{fig:iran_us_dw}
  \end{subfigure}

  \caption{Screenshots for the videos in Type 5, 6, 8, 9, and 10.}
  \label{fig:type5}
\end{figure}

\begin{itemize}[leftmargin=1.2em, itemsep=2pt, topsep=2pt, parsep=0pt]

  \item Bloomberg News: \url{https://www.youtube.com/watch?v=PRgLJcVr1DM}
 \item The Economic Times:  \url{https://www.youtube.com/watch?v=U8yD0JNCWrw}

  \item Breaking Now:  \url{https://www.youtube.com/watch?v=rlM-nprMIdg}

  \item Global News: \url{https://www.youtube.com/watch?v=0pH2-pppbSQ}

\end{itemize}

\textbf{Multiple Choices:}
A. Bloomberg News uses a studio anchor with minimal visuals to emphasize the political win for Biden, whereas Breaking Now employs direct footage of leaders and symbolic b-roll to frame the deal as a humanitarian emergency.\\
B. The Economic Times overlays an on-screen graphic about the loan amount while showing Biden speaking, while Global News uses stock footage of urban buildings to imply economic consequences without narrating them. \\
C. Breaking Now pairs its narration about legal constraints with a blood-stained handshake and a destroyed tank, while Bloomberg News supports its narrative on Biden's political triumph with field footage of leaders walking and shaking hands. \\
D. Global News uses a talking-head format with a static backdrop to discuss the Canadian initiative, whereas Bloomberg News contrasts studio commentary with clips of Biden and Zelensky to convey urgency.

\textbf{Correct choice:} C

\textbf{Explanation:}
Breaking Now: A destroyed tank with graffiti is shown while narrating legal challenges to asset use and a blood-stained handshake symbolizes conflict while explaining the deal's urgency. Bloomberg News: President Biden walks toward the summit. Bloomberg News: Biden and Zelensky shake hands after the agreement. Anchor Anamalie Hordern discusses Biden's political benefit from the deal. Global News: talking-head format, no static backdorop.

\hrulefill

\subsection{Layer 2: Cross-source News Integration}
\item{Type 6: Cross-source cross-modal evidence integration.} This type of questions test a model's understanding across news reportings from multiple sources (news outlets) in integrating their audio(textual)-video evidences into a reasoning chain, in the meantime resolving their differences and filtering out evidences that conflict one another. It also checks whether evidences from one reporting can be used to support the claims in another reporting.  

\textbf{Sample Question:}
Which statement accurately combines textual and visual evidence from multiple sources regarding the G7's plan to support Ukraine using frozen Russian assets? (Screenshots are shown in Figure\ref{fig:type5})

\begin{itemize}[leftmargin=1.2em, itemsep=2pt, topsep=2pt, parsep=0pt]

  \item Bloomberg News: \url{https://www.youtube.com/watch?v=PRgLJcVr1DM}
 \item The Economic Times:  \url{https://www.youtube.com/watch?v=U8yD0JNCWrw}

  \item Breaking Now:  \url{https://www.youtube.com/watch?v=rlM-nprMIdg}

  \item Global News: \url{https://www.youtube.com/watch?v=0pH2-pppbSQ}

\end{itemize}

\textbf{Multiple Choices:}
A. The Economic Times states the G7 signed a plan to unlock \$50 billion from frozen assets while Global News visually shows Justin Trudeau walking with other G7 leaders, confirming his participation in the agreement.\\
B. Breaking Now mentions the loan will be mostly guaranteed by the US government and Global News shows a tank in Ukraine, suggesting military use of the loan, but the frame does not confirm this use. \\
C. Bloomberg News claims the plan is a big win for the Biden administration and Breaking Now visually confirms the \$50 billion figure in a text overlay during the discussion of the deal. \\
D. Global News reports Canada is backing \$15 billion of the loan and The Economic Times visually shows President Biden at a podium speaking about the agreement, reinforcing his leadership role.

\textbf{Correct choice:} A

\textbf{Explanation:}
The Economic Times states the G7 signed a plan to unlock \$50 billion from frozen assets, and Global News shows Canada is backing \$5 billion of the loan and Justin Trudeau walking with other G7 leaders, confirming his participation in the agreement.

\hrulefill

\item{Type 7: Cross-source evidence-claim conflict detection.} This type of questions test a model's understanding across news reportings from multiple sources in detecting an evidence from one reporting that contradicts to an claim in another reporting.

\textbf{Sample Question:}
Which claim about the refugee situation is contradicted by cross-source evidence from the Associated Press and Bloomberg News?
(Screenshots are shown in Figure\ref{fig:type7})

\begin{figure}[H]
  \centering

  \begin{subfigure}[t]{0.32\textwidth}
    \centering
    \includegraphics[width=\linewidth]{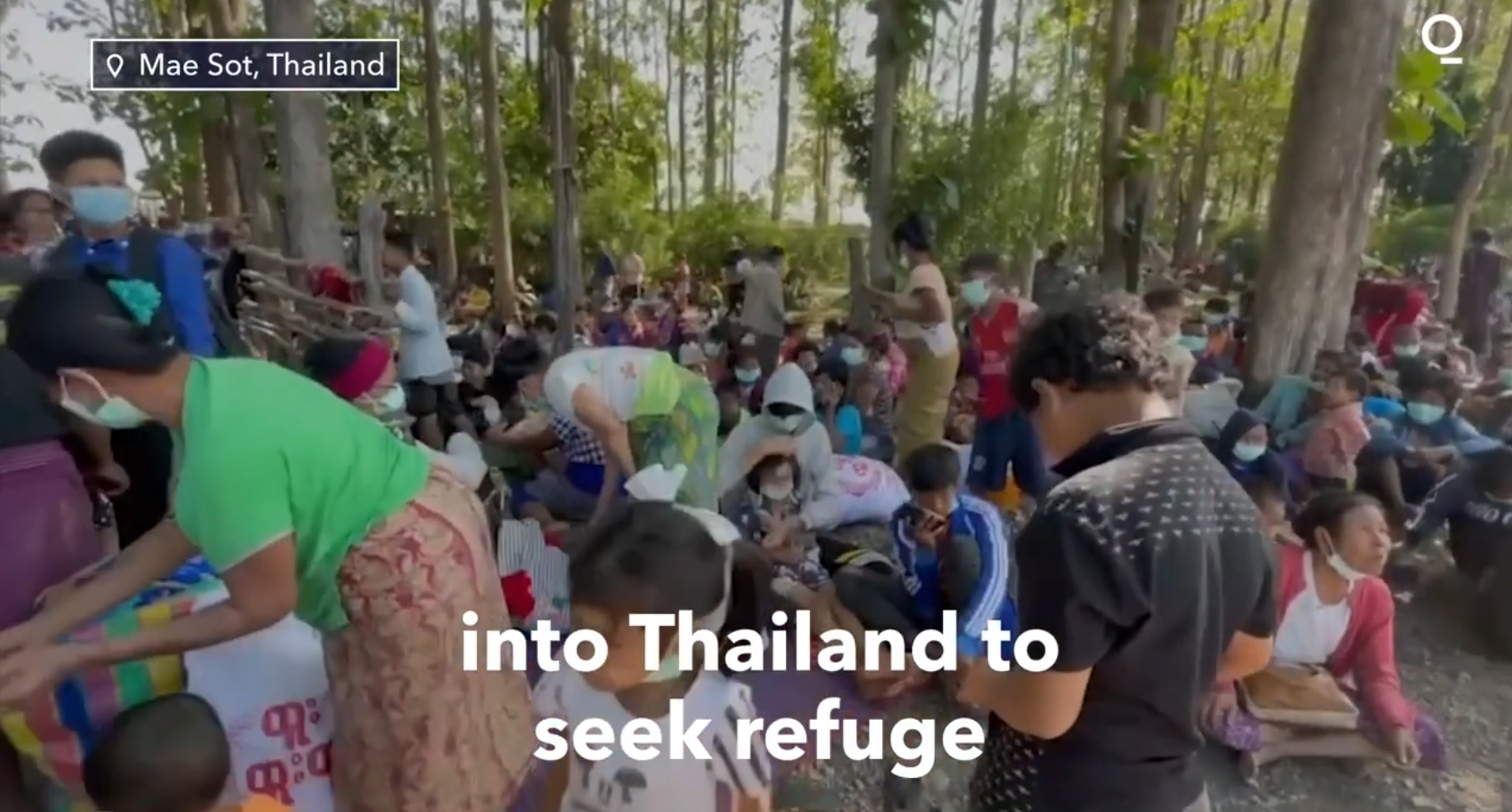}
    \caption{Bloomberg News}
    \label{fig:iran_us_nbc}
  \end{subfigure}
  \hfill
  \begin{subfigure}[t]{0.32\textwidth}
    \centering
    \includegraphics[width=\linewidth]{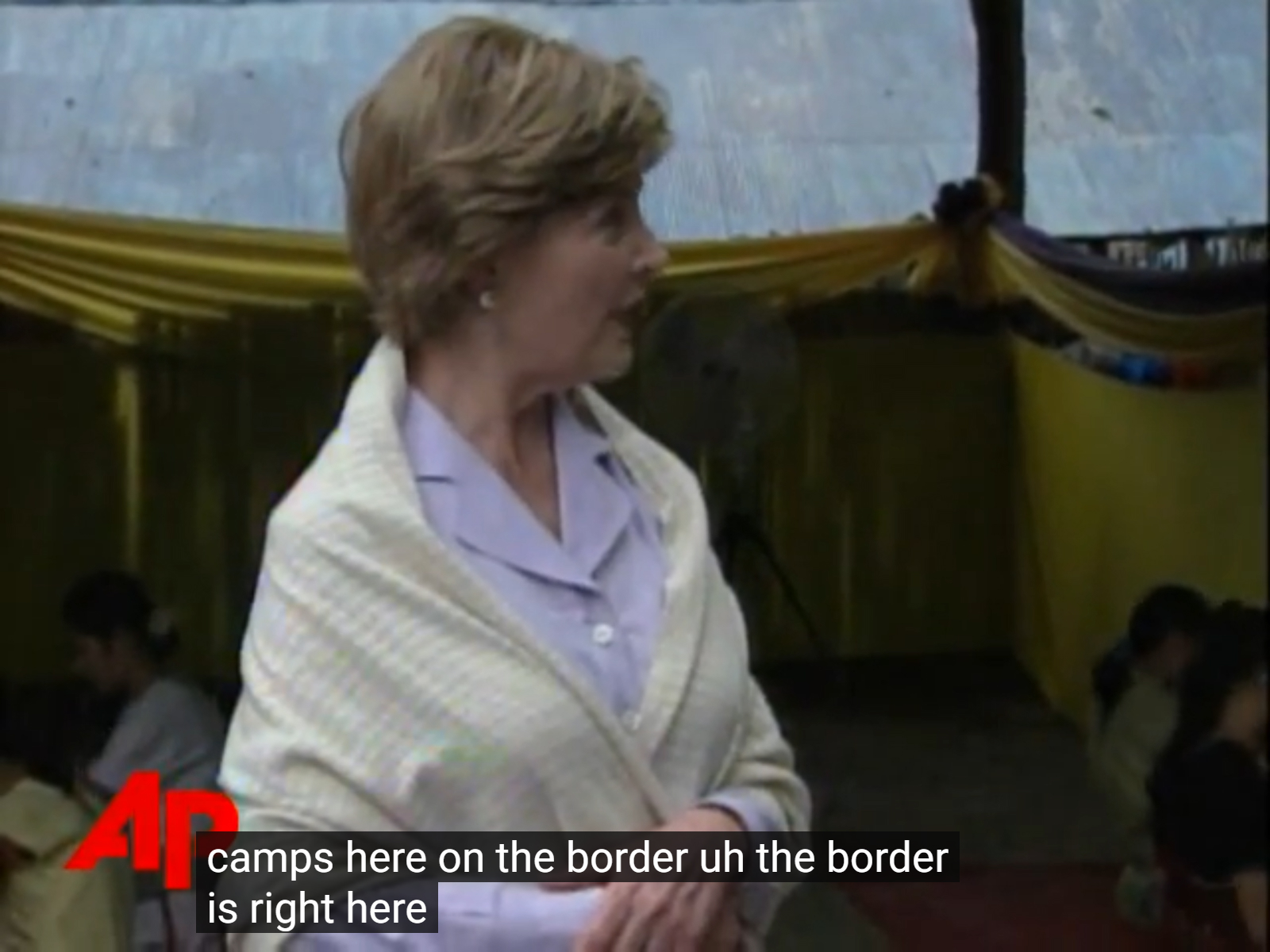}
    \caption{Associated Press}
    \label{fig:iran_us_euronews}
  \end{subfigure}

  \caption{Screenshots for the videos in Type 7.}
  \label{fig:type7}
\end{figure}

\begin{itemize}[leftmargin=1.2em, itemsep=2pt, topsep=2pt, parsep=0pt]

  \item Associated Press: \url{https://www.youtube.com/watch?v=QeSiaZy1ksI}
 \item Bloomberg News:  \url{https://www.youtube.com/watch?v=veeDpBn30r8}

\end{itemize}

\textbf{Multiple Choices:}
A. The Associated Press claims the Thai government is generous, while Bloomberg’s text overlay labels the area as Mae Sot, Thailand, which does not confirm government generosity.\\
B. The Associated Press claims the camp is a place of education, and Bloomberg’s frame shows a fire in Mae Sot, which contradicts any claim of peaceful conditions. \\
C. The Associated Press claims the refugees are engaged in daily life, while Bloomberg shows ethnic Karen guerrillas, indicating the refugees are not just civilians.\\
D. The Associated Press claims the camp is on the Thai-Burma border, but Bloomberg’s footage shows people in Mae Sot, Thailand, which is not the same location.
    
\textbf{Correct choice:} D

\textbf{Explanation:}

Associated Press shows a camp with 'AP' logo, and its transcript describes the location as 'the border with Burma'. Bloomberg shows a location label 'Mae Sot, Thailand', which is a town along the border — but the AP transcript calls it 'Camp' and does not claim Mae Sot as the camp's name, suggesting different physical or administrative locations.

\hrulefill

\item{Type 8: Cross-source temporal development.} This type of questions test a model's understanding across news reportings from multiple sources in establishing the temporal order of the reportings and the temporal development of a story.  

\textbf{Sample Question:}
Which sequence correctly reconstructs the real-world temporal development of the G7's decision on using frozen Russian assets for Ukraine, based on all sources? (Screenshots are shown in Figure\ref{fig:type5})

\begin{itemize}[leftmargin=1.2em, itemsep=2pt, topsep=2pt, parsep=0pt]

  \item Bloomberg News: \url{https://www.youtube.com/watch?v=PRgLJcVr1DM}
 \item The Economic Times:  \url{https://www.youtube.com/watch?v=U8yD0JNCWrw}

  \item Breaking Now:  \url{https://www.youtube.com/watch?v=rlM-nprMIdg}

  \item Global News: \url{https://www.youtube.com/watch?v=0pH2-pppbSQ}

\end{itemize}

\textbf{Multiple Choices:}
A. President Biden called for immediate use of frozen Russian assets to aid Ukraine in 2023, and by June 2024, G7 leaders agreed to a \$50 billion loan from the interest on those funds, with the first meeting to discuss the plan held in Bari, Italy.\\
B. The G7 frozen \$280 billion in Russian assets in 2022 after Russia's invasion, and later agreed in June 2024 to loan \$50 billion from the interest earned on those funds to Ukraine, with President Biden stating the goal is to provide immediate resources for Ukraine's economic and energy needs.\\
C. G7 leaders signed a \$50 billion loan agreement in June 2024, using profits from frozen Russian assets, following President Biden's announcement that the deal was finally getting over the finish line, with Canada's role in the agreement being debated prior to the summit. \\
D. Before the G7 summit in June 2024, leaders of the world's advanced economies had already discussed a \$69 billion aid package for Ukraine with \$50 billion coming from Russian assets, and then officially agreed during the summit to a \$50 billion loan from the interest of the frozen capital, which was first frozen by the G7 in 2022.

\textbf{Correct choice:} B

\textbf{Explanation:}
Breaking Now: the G7 froze \$280 billion in Russian assets in 2022 after Russia's invasion; The Economic Times: the G7 agreed in June 2024 to loan \$50 billion from the interest on those funds to Ukraine.

\hrulefill

\item{Type 9: Cross-source narrative reconstruction.} This type of questions test a model's understanding across news reportings from multiple sources by integrating their details into one comprehensive and coherence story.  

\textbf{Sample Question:}
Which of the following correctly reconstructs the full narrative arc of the G7's new financial support for Ukraine, based on the combined coverage from Bloomberg News, The Economic Times, Breaking Now, and Global News? (Screenshots are shown in Figure\ref{fig:type5})

\begin{itemize}[leftmargin=1.2em, itemsep=2pt, topsep=2pt, parsep=0pt]

  \item Bloomberg News: \url{https://www.youtube.com/watch?v=PRgLJcVr1DM}
 \item The Economic Times:  \url{https://www.youtube.com/watch?v=U8yD0JNCWrw}

  \item Breaking Now:  \url{https://www.youtube.com/watch?v=rlM-nprMIdg}

  \item Global News: \url{https://www.youtube.com/watch?v=0pH2-pppbSQ}

\end{itemize}

\textbf{Multiple Choices:}
A. Bloomberg News provides the initial trigger by emphasizing Biden's press conference and the strategic significance of the agreement, while The Economic Times describes the escalation through the announcement of the \$50 billion loan and the broader geopolitical framing of countering Russia's aggression. \\
B. Breaking Now covers the policy response by detailing how the loan will be structured through profits earned on frozen assets, while Global News explains the aftermath by showing the potential impact on international leaders and diplomatic relationships.\\
C. Global News establishes the cause with images of Russian assets and economic centers, while Breaking Now reconstructs the escalation with visuals of conflict on the ground and the subsequent decision to use frozen assets for financing. \\
D. The Economic Times frames the narrative with the immediate aftermath and the moral justification for the loan, while Bloomberg News adds the policy response with details on implementation and bilateral agreements signed at the summit.

\textbf{Correct choice:} A

\textbf{Explanation:}
Bloomberg News: no comprehensive details on implementation of \$50 billion loan; Global News: no potential impacts on international leaders, Russian spots presented, instead of Russian assets.

\hrulefill

\item{Type 10: Multi-source summary.} This type of questions test a model's understanding across news reportings from multiple sources by making a comprehensive summary, including what all of them can agree on and what they disagree. 

\textbf{Sample Question:}
Which summary best captures the cross-source understanding of the G7's decision on Ukrainian aid, as reported by Bloomberg News, The Economic Times, Breaking Now, and Global News? (Screenshots are shown in Figure\ref{fig:type5})

\begin{itemize}[leftmargin=1.2em, itemsep=2pt, topsep=2pt, parsep=0pt]

  \item Bloomberg News: \url{https://www.youtube.com/watch?v=PRgLJcVr1DM}
 \item The Economic Times:  \url{https://www.youtube.com/watch?v=U8yD0JNCWrw}

  \item Breaking Now:  \url{https://www.youtube.com/watch?v=rlM-nprMIdg}

  \item Global News: \url{https://www.youtube.com/watch?v=0pH2-pppbSQ}

\end{itemize}

\textbf{Multiple Choices:}
A. The G7 reached a consensus to lend 50 billion to Ukraine using profits from frozen Russian assets, with all sources agreeing on the financial mechanism and the goal of supporting Ukraine's resilience against Russian aggression. \\
B. There is disagreement among the G7 nations about the legality of using frozen Russian assets for Ukraine's aid, with some outlets highlighting the 50 billion loan while others note that the full confiscation of assets remains controversial and unconfirmed. \\
C. All outlets report that the G7 approved a 50 billion loan for Ukraine, but they differ on the financial specifics and the political implications, with one source suggesting the decision is driven by the upcoming US elections. \\
D. The primary dispute across the reports is whether the loan should be based on the freezing of Russian assets, with some sources asserting it is a win for the Biden administration while others claim the funds will not be distributed immediately due to legal hurdles.

\textbf{Correct choice:} A

\textbf{Explanation:}
All sources support the consensus to lend 50 billion to Ukraine using profits from frozen Russian assets.

\end{itemize}

\section{QA generation}\label{apd:qa_gen}


To generate drafts of questions, we applied multi-level guidance prompts. The prompt contains three parts, type taxonomy, type head, and common rules. The three parts together instruct the MLLM understand the key points of each type, what the question should and should no focus on. Here is the prompt:

\begin{promptbox}{QA Generation}
\raggedright
\sloppy
\textbf{System Prompt:}
\begin{itemize}[leftmargin=1.2em, itemsep=2pt, topsep=2pt, parsep=0pt]
  \item type taxonomy
  \item type head
  \item common rules
\end{itemize}

\textbf{Human Prompt:}
\begin{itemize}[leftmargin=1.2em, itemsep=2pt, topsep=2pt, parsep=0pt]
  \item \textbf{Transcripts:} \newline [TAG\_$1$]: <transcript\_$1$>\newline[TAG\_$2$]: <transcript $2$>\newline...\newline[TAG\_$N$]: <transcript\_$N$>

  \item \textbf{Videos:}  \newline[TAG\_$1$]:<Frame $1$><Frame\_$2$>...<Frame\_$M$>\newline[TAG\_$2$]:<Frame $1$><Frame\_$2$>...<Frame\_$M$>\newline...\newline[TAG\_$N$]:<Frame\_$1$><Frame\_$2$>...<Frame\_$M$>

\end{itemize}
\end{promptbox}

\subsection{Type Taxonomy}
\begin{promptbox}{Type Taxonomy}
\begin{itemize}[leftmargin=1.2em, itemsep=4pt, topsep=4pt]

  \item All questions in this benchmark belong to one of TEN predefined question types (Type~1--Type~10).
  \item Each type targets a distinct reasoning skill.
  \item It is critical that each question strictly matches its assigned type and does NOT overlap with the reasoning focus of other types.

  \item The ten question types are organized into two groups:

  \item \textbf{(A) Multi-source News Comparison (Types 1--5)}  
  These types focus on comparing how different outlets report the SAME story.
  
  \begin{itemize}[leftmargin=1.4em, itemsep=3pt]
    \item \textbf{Type 1: Main Claim Comparison}  
    Compare the overarching claims, conclusions, or stances taken by different outlets.  
    Focus on opinions, interpretations, or judgments, not fine-grained details.

    \item \textbf{Type 2: Event Comparison}  
    Compare concrete, verifiable details across outlets, such as numbers, locations, dates, entities, or explicitly stated causes.

    \item \textbf{Type 3: Cross-video Visual Comparison}  
    Compare what is visually shown across different outlets’ videos.  
    Focus on visible elements in frames, not what is said in transcripts.

    \item \textbf{Type 4: Perspective / Framing Comparison}  
    Compare narrative perspectives, interpretive lenses, or emphasis.  
    Focus on how the story is framed, not on raw events or visuals alone.

    \item \textbf{Type 5: Multi-modal Presentation Comparison}  
    Compare how narration, visuals, and on-screen graphics are combined to present information.  
    Focus on presentation strategy rather than content alone.
  \end{itemize}

  \item \textbf{(B) Cross-source News Integration (Types 6--10)}  
  These types require integrating information across multiple sources and modalities.

  \begin{itemize}[leftmargin=1.4em, itemsep=3pt]
    \item \textbf{Type 6: Cross-source Cross-modal Evidence Integration}  
    Integrate textual/audio evidence from one outlet with visual evidence from another.  
    No single source or modality is sufficient on its own.

    \item \textbf{Type 7: Cross-source Evidence--Claim Conflict Detection}  
    Identify when evidence from one outlet contradicts claims made by another.

    \item \textbf{Type 8: Cross-source Temporal Development}  
    Reconstruct the real-world timeline of events by combining reports from different outlets.  
    This is about event chronology, not reporting order.

    \item \textbf{Type 9: Cross-source Narrative Reconstruction}  
    Reconstruct a coherent multi-stage story by combining complementary narrative segments from different outlets.

    \item \textbf{Type 10: Multi-source Summary}  
    Produce a high-level synthesis that captures both shared events and key differences across all outlets.
  \end{itemize}

  \item \textbf{IMPORTANT:}
  \begin{itemize}[leftmargin=1.4em, itemsep=3pt]
    \item Each question must belong to \textbf{EXACTLY ONE} type.
    \item The reasoning focus of the assigned type must be clearly identifiable.
    \item Do \textbf{NOT} mix reasoning goals from multiple types in a single question.
  \end{itemize}

\end{itemize}
\end{promptbox}

\subsection{Type heads}
\begin{promptbox}{Type 1: Main Claim Comparison}
\textbf{Goal.}
Generate \textbf{one hard multiple-choice question (MCQ)} that evaluates a model’s ability to
compare the \textbf{main claims or overall conclusions} made by different news sources
about the same underlying story.

\begin{itemize}[leftmargin=1.2em, itemsep=2pt, topsep=2pt, parsep=0pt]

  \item You will receive raw transcripts with explicit source tags and a small number of frames.
  \item The question must test understanding of how different sources
  \textbf{frame, interpret, or conclude} the shared story, rather than recalling isolated events.

\end{itemize}

\textbf{Core concept: Main claim.}
A main claim is the headline-level conclusion or organizing message of a report, such as:
\begin{itemize}[leftmargin=1.2em, itemsep=1pt, topsep=1pt, parsep=0pt]
  \item what is happening overall,
  \item why it matters,
  \item what the primary cause, risk, or responsibility is,
  \item what response is implied or recommended.
\end{itemize}
It is \textbf{not} a single number, minor detail, or isolated statement.

\textbf{What the question must focus on:}
\begin{itemize}[leftmargin=1.2em, itemsep=2pt, topsep=2pt, parsep=0pt]
  \item The central claim, verdict, or overarching conclusion each source presents.
  \item Similarities and differences among these main claims
  (e.g., systemic crisis vs.\ limited impact, economic vs.\ political framing,
  stabilizing vs.\ deteriorating outlook).
  \item Reasoning that requires comparing \textbf{at least two sources}, not restating a single report.
\end{itemize}

\textbf{What the question must not focus on:}
\begin{itemize}[leftmargin=1.2em, itemsep=2pt, topsep=2pt, parsep=0pt]
  \item Fine-grained event details (exact numbers, minor entities, technical percentages).
  \item Purely visual layout or presentation style (these belong to other task types).
  \item Explicit evidence--claim contradiction detection (Type~7).
  \item Pure temporal development, narrative reconstruction, or global summaries
  (Types~8 and~10).
\end{itemize}

\textbf{Use of transcripts and frames:}
\begin{itemize}[leftmargin=1.2em, itemsep=2pt, topsep=2pt, parsep=0pt]
  \item Transcripts and visual cues may be used to infer each source’s main claim.
  \item The stem must remain a comparison of \textbf{claims or conclusions},
  not a question about specific images or on-screen text.
  \item Visual cues may inform claim interpretation, but the question must not
  degenerate into a Type~3 or Type~5 task.
\end{itemize}

\textbf{Representative example (style only; do not reuse content):}
A valid Type~1 question explicitly compares how different sources frame the same story
at the level of overarching conclusions, with one option correctly capturing the
cross-source contrast and all others remaining plausible but incorrect under joint consideration.

\end{promptbox}

\begin{promptbox}{Type 2: Event Comparison}

\textbf{Goal.}
Generate \textbf{one hard multiple-choice question (MCQ)} that evaluates a model’s ability
to compare and reconcile \textbf{details of event} reported by different news sources
about the same underlying story.

\begin{itemize}[leftmargin=1.2em, itemsep=2pt, topsep=2pt, parsep=0pt]
  \item You will receive raw transcripts with explicit source tags and a small number of frames.
  \item The question must test cross-source reasoning over \textbf{concrete, verifiable details},
  rather than tone, framing, or narrative style.
\end{itemize}

\textbf{Core concept: Details of events.}
The details of an event refers to concrete, checkable information, including:
\begin{itemize}[leftmargin=1.2em, itemsep=1pt, topsep=1pt, parsep=0pt]
  \item numbers, counts, proportions, or percentages;
  \item specific locations or named entities;
  \item clearly stated causes, triggers, or outcomes;
  \item explicit timelines or dated events;
  \item direct quotations of key actors asserting verifiable claims.
\end{itemize}
The focus is on \textbf{what is reported as true} based on the details, not on interpretation or presentation.

\textbf{What the question must focus on:}
\begin{itemize}[leftmargin=1.2em, itemsep=2pt, topsep=2pt, parsep=0pt]
  \item How the event compares across sources:
  alignment, divergence, omission, or refinement.
  \item Whether reported events are consistent, complementary, or conflicting.
  \item Reasoning that requires examining events from
  \textbf{at least two distinct sources}.
\end{itemize}

\textbf{What the question must not focus on:}
\begin{itemize}[leftmargin=1.2em, itemsep=2pt, topsep=2pt, parsep=0pt]
  \item Explicit references to images, footage, screenshots, or on-screen layout.
  \item Questions about what is visually shown or which frame contains a graphic
  (these belong to visual or multi-modal task types).
  \item Narrative framing, tone, or stylistic differences.
  \item Evidence--claim contradiction detection, which belongs to Type~7.
\end{itemize}

\textbf{Use of image frames (internal only):}
\begin{itemize}[leftmargin=1.2em, itemsep=2pt, topsep=2pt, parsep=0pt]
  \item Frame content may be used to verify or disambiguate details
  (e.g., a number on a chart or a place name on a map).
  \item Evidence may cite frame-based details using standard
  \texttt{[FRAME: TAG\_X \#k]} markers, while the question wording remains non-visual.
\end{itemize}

\textbf{Structural requirements for the MCQ:}
\begin{itemize}[leftmargin=1.2em, itemsep=2pt, topsep=2pt, parsep=0pt]
  \item The stem must clearly indicate that the question concerns
  \textbf{reported details} across sources.
  \item Each option (A--D) must combine \textbf{two short sub-claims},
  for example, one clause describing what TAG\_A reports and one clause
  describing what TAG\_B or TAG\_C reports.
  \item \textbf{Exactly one} option must correctly capture the cross-source
   relationship when all sources are considered together.
  \item Each distractor must be locally plausible but must distort, omit,
  or miscombine evidence so that it becomes incorrect under full comparison.
\end{itemize}

\textbf{Representative example (style only; do not reuse content):}
A valid Type~2 question compares concrete details such as numbers, locations,
or timelines across sources, with one option correctly integrating the
reported evidence and all others remaining plausible but incorrect under
careful cross-source verification.

\end{promptbox}

\begin{promptbox}{Type 3: Visual Comparison}

\textbf{Goal.}
Generate \textbf{one hard multiple-choice question (MCQ)} that evaluates a solver’s ability
to compare \textbf{visual evidence across videos} from different news sources
covering the same underlying story.

\begin{itemize}[leftmargin=1.2em, itemsep=2pt, topsep=2pt, parsep=0pt]
  \item You will receive raw transcripts with explicit source tags and a set of extracted frames from each source.
  \item The question must require reasoning based on \textbf{what appears on screen},
  rather than on transcript content.
\end{itemize}

\textbf{Core concept: Visual comparison.}
Type~3 evaluates a model’s capacity to:
\begin{itemize}[leftmargin=1.2em, itemsep=1pt, topsep=1pt, parsep=0pt]
  \item identify what appears visually in each source’s footage;
  \item track where and when specific visuals appear
  (e.g., opening, mid-segment, closing, or insert);
  \item detect presence, absence, reuse, or substitution of imagery across sources;
  \item compare visual styles (studio vs.\ field footage, maps vs.\ B-roll,
  graphics vs.\ raw scenes);
  \item recognize fine-grained visual attributes, including:
  people, actions, objects, colors, clothing, environments, signage,
  locations, graphical elements, charts, maps, lower-thirds,
  CCTV, drone, on-site, archival, or animated sequences.
\end{itemize}

The question must \textbf{hinge on visual inspection}, not on narration or textual description.

\textbf{What the question must focus on:}
\begin{itemize}[leftmargin=1.2em, itemsep=2pt, topsep=2pt, parsep=0pt]
  \item Cross-video distribution of visual elements:
  which sources show them, which do not, and when they appear.
  \item Subtle but consequential visual differences, such as:
  maps versus graphics, CCTV versus phone footage,
  nighttime versus daytime scenes, or field shots versus studio presentation.
  \item Visual-only cues that cannot be inferred from transcripts alone.
\end{itemize}

\textbf{What the question must not become:}
\begin{itemize}[leftmargin=1.2em, itemsep=2pt, topsep=2pt, parsep=0pt]
  \item Pure numerical comparison derived from text (Type~2).
  \item Framing or perspective comparison (Type~4).
  \item Multi-modal integration or presentation style comparison (Type~5).
  \item Any question answerable without inspecting the visuals.
  \item Evidence--claim conflict detection across sources, which belongs to Type~7.
\end{itemize}

\textbf{Structural requirements for the MCQ:}
\begin{itemize}[leftmargin=1.2em, itemsep=2pt, topsep=2pt, parsep=0pt]
  \item Use \textbf{at least two distinct} source tags from the provided set.
  \item The question text must explicitly mention all chosen tags.
  \item Each option (A--D) must be a concise comparison of visual details,
  combining which source shows a visual element and how or where
  another source shows (or does not show) it.
  \item \textbf{Exactly one} option must be correct.
\end{itemize}

\textbf{Use of frames (mandatory):}
\begin{itemize}[leftmargin=1.2em, itemsep=2pt, topsep=2pt, parsep=0pt]
  \item The correct answer must rely on \textbf{specific, observable visual evidence}.
  \item Acceptable visual cues include:
  CCTV footage, animated maps, on-screen graphics,
  time-of-day differences, studio versus field shots,
  or distinctive people, objects, or locations.
\end{itemize}

\textbf{Representative examples (style only; do not reuse content):}
A valid Type~3 question compares the distribution, timing, or form of visual elements
across sources, with one option accurately reflecting the visual evidence and all others
remaining plausible but incorrect under careful cross-video inspection.

\end{promptbox}

\begin{promptbox}{Type 4: Narrative Angle Comparison}
\textbf{Goal.}
Generate \textbf{one hard multiple-choice question (MCQ)} that evaluates how well a solver can compare
\textbf{distinct perspectives, interpretive lenses, or narrative angles} adopted by different sources
when presenting the same underlying event.

\textbf{Core concept: Perspective / interpretive lens / narrative angle.}
This refers to how a source interprets, positions, or contextualizes the story, including:
\begin{itemize}[leftmargin=1.2em, itemsep=1pt, topsep=1pt, parsep=0pt]
  \item which actors, causes, or consequences it emphasizes or downplays;
  \item whether the language is critical, sympathetic, neutral, technocratic, alarmed, etc.;
  \item what values or priorities it highlights (e.g., public safety, markets, victims, governance);
  \item what storyline it foregrounds (e.g., policy success, institutional risk, human impact, geopolitical tension).
\end{itemize}
This is about \textbf{how the story is interpreted}, not merely \textbf{what happened}.

\textbf{What the question must focus on:}
\begin{itemize}[leftmargin=1.2em, itemsep=2pt, topsep=2pt, parsep=0pt]
  \item Similarities and differences in interpretive perspective across sources
  (e.g., technical adjustment vs.\ crisis framing; victim-centered vs.\ institutional lens; blame assignment vs.\ structural complexity).
  \item Inference of interpretive angles primarily from wording, tone, and emphasis, and secondarily from supportive visuals when relevant.
\end{itemize}

\textbf{What the question must not become:}
\begin{itemize}[leftmargin=1.2em, itemsep=2pt, topsep=2pt, parsep=0pt]
  \item Pure events' details comparison without interpretive distinctions (Type~2).
  \item Pure visual-only comparison of what is shown (Type~3).
  \item Multi-modal presentation mechanics comparison (Type~5).
  \item Evidence--claim conflict detection across sources (Type~7).
  \item Timeline or reconstruction questions (Types~8--9).
\end{itemize}

\textbf{Use of transcripts and frames:}
\begin{itemize}[leftmargin=1.2em, itemsep=2pt, topsep=2pt, parsep=0pt]
  \item Detect perspective primarily from wording, tone, and emphasis (e.g., evaluative adjectives, quote selection, framing of drivers/risks/significance).
  \item Visuals may reinforce perspective (e.g., officials vs.\ crowds, data charts vs.\ human scenes), but the comparison remains interpretive rather than purely visual.
\end{itemize}

\textbf{Structural requirements for the MCQ:}
\begin{itemize}[leftmargin=1.2em, itemsep=2pt, topsep=2pt, parsep=0pt]
  \item The stem must clearly indicate that the type concerns \textbf{how sources interpret or position the story}.
  \item Each option (A--D) must combine \textbf{two short sub-claims} about perspectives
  (e.g., TAG\_A’s angle and TAG\_B’s/TAG\_C’s angle) in a single sentence.
  \item \textbf{Exactly one} option must correctly capture the cross-source pattern of perspectives.
  \item Distractors must be plausible but incorrect by misattribution, exaggeration/flattening, or incompatible mixing of angles.
\end{itemize}

\end{promptbox}

\begin{promptbox}{Type 5: Multi-modal Presentation Comparison}

\textbf{Goal.}
Generate \textbf{one hard multiple-choice question (MCQ)}  that evaluates how well a solver can compare
\textbf{how different sources integrate narration and visuals} to present evidence and arguments.

\textbf{Core concept: Multi-modal presentation.}
Multi-modal presentation refers to how a source combines:
spoken narration, on-screen text (lower-thirds, tickers, captions),
graphics (charts, infographics, maps), and raw footage/B-roll (field scenes, interviews, events),
with emphasis on \textbf{how these modalities are coordinated} to convey the story.

\textbf{What the question must focus on:}
\begin{itemize}[leftmargin=1.2em, itemsep=2pt, topsep=2pt, parsep=0pt]
  \item Cross-source differences in integration strategy (e.g., chart-heavy with verbal explanation vs.\ anchor-led with minimal visuals; organized alternation of interviews/B-roll/overlays).
  \item Alignment between narration and visuals (e.g., visuals directly illustrate spoken points vs.\ visuals remain generic while narration becomes specific).
\end{itemize}

\textbf{What the question must not reduce to:}
\begin{itemize}[leftmargin=1.2em, itemsep=2pt, topsep=2pt, parsep=0pt]
  \item Pure details of event comparison with no presentation focus (Type~2).
  \item Pure visual-only comparison without narration/audio context (Type~3).
  \item Pure perspective comparison without modality integration (Type~4).
  \item Pure timeline or narrative-phase questions (Types~8--9).
\end{itemize}

\textbf{Use of transcripts and frames:}
\begin{itemize}[leftmargin=1.2em, itemsep=2pt, topsep=2pt, parsep=0pt]
  \item Use transcripts to infer narration detail, references to graphics, and switching patterns (studio vs.\ interviews/field).
  \item Use frames to identify visual elements (charts, maps, tickers, full-screen graphics, field scenes) and whether they match narration at that moment.
  \item The correct option must require attending to \textbf{both} narration and visuals; it must not be solvable from either modality alone.
\end{itemize}

\textbf{Structural requirements for the MCQ:}
\begin{itemize}[leftmargin=1.2em, itemsep=2pt, topsep=2pt, parsep=0pt]
  \item The stem must explicitly ask about how sources \textbf{use and combine} narration and visuals.
  \item Each option (A--D) must combine two short sub-claims (e.g., TAG\_A’s integration pattern and TAG\_B’s/TAG\_C’s contrasting pattern).
  \item \textbf{Exactly one} option must correctly describe the cross-source multi-modal pattern under joint evidence.
  \item Distractors may sound plausible from one modality alone but fail when both narration and visuals are checked.
\end{itemize}

\end{promptbox}

\begin{promptbox}{Type 6: Cross-source Cross-modal Evidence Integration}

\textbf{Goal.}
Generate \textbf{one hard multiple-choice question (MCQ)} that forces the solver to \textbf{integrate evidence across multiple sources and multiple modalities}
(text/audio and visuals) to reach a single well-supported conclusion.

\textbf{Core concept: Cross-source cross-modal integration.}
The correct answer must require combining textual/spoken information from multiple sources and
visual evidence from frames (on-screen text, maps, labels, charts, visible actors/scenes).
A single source or single modality must be insufficient.

\textbf{What the question must focus on:}
\begin{itemize}[leftmargin=1.2em, itemsep=2pt, topsep=2pt, parsep=0pt]
  \item Integration patterns such as: visual cues from one source paired with textual claims from another; resolving mismatched numbers/locations; visuals confirming or undermining narration; merging partial accounts into a complete conclusion.
  \item Reasoning that depends on \textbf{multi-source and multi-modal} evidence; if solvable from one transcript alone, it is not Type~6.
\end{itemize}

\textbf{What the question must not reduce to:}
\begin{itemize}[leftmargin=1.2em, itemsep=2pt, topsep=2pt, parsep=0pt]
  \item Pure details of events comparison (Type~2) or pure visual comparison (Type~3).
  \item Pure framing comparison (Type~4) or single-source narration--visual matching (Type~5).
  \item Temporal development or narrative reconstruction (Types~8--9).
  \item High-level multi-source summaries (Type~10).
\end{itemize}

\textbf{Use of transcripts and frames:}
\begin{itemize}[leftmargin=1.2em, itemsep=2pt, topsep=2pt, parsep=0pt]
  \item Use transcripts to extract claims, numbers, actors, causes, implications, or verbal contradictions.
  \item Use frames to extract decisive on-screen numbers/labels/maps/banners and to test narration--visual consistency.
  \item The correct option must rely on at least one decisive visual cue and at least one decisive textual/audio cue from \textbf{different sources}.
\end{itemize}

\textbf{Structural requirements for the MCQ:}
\begin{itemize}[leftmargin=1.2em, itemsep=2pt, topsep=2pt, parsep=0pt]
  \item The stem must signal that answering requires piecing together evidence from multiple sources.
  \item Each option (A--D) must combine two short sub-claims that reflect cross-modal integration across sources.
  \item \textbf{Exactly one} option must match the fully integrated reasoning chain.
  \item Distractors must fail by ignoring a modality, relying on a single source, mishandling contradictions, or combining misaligned details.
\end{itemize}

\end{promptbox}

\begin{promptbox}{Type 7: Cross-source Evidence--Claim Conflict Detection}

\textbf{Goal.}
Generate \textbf{one hard multiple-choice question (MCQ)} that evaluates whether a solver can detect
\textbf{when evidence from one source contradicts or undermines a claim made in another source}.

\textbf{Core concept: Evidence--claim conflict.}
Type~7 targets cases where one source makes an explicit claim (numbers, causes, locations, outcomes),
and another source provides event evidence (transcript details and/or visuals) that contradicts it
or clearly fails to support it.

\textbf{What the question must focus on:}
\begin{itemize}[leftmargin=1.2em, itemsep=2pt, topsep=2pt, parsep=0pt]
  \item Identify a specific explicit claim in one source.
  \item Identify evidence in other source(s) that directly contradicts it or shows it is incomplete/misleading.
  \item Ask which cross-source combination reveals the conflict, or which claim cannot be maintained under joint evidence.
\end{itemize}

\textbf{Optional secondary pattern (allowed but not required):}
Evidence from one source may strongly verify a controversial claim made in another source while others remain neutral.
There must still be a clear supported-versus-contradicted contrast.

\textbf{What the question must not reduce to:}
\begin{itemize}[leftmargin=1.2em, itemsep=2pt, topsep=2pt, parsep=0pt]
  \item Simple events alignment/mismatch without explicit claim versus evidence structure (Type~2).
  \item Pure visual comparison without explicit claim reference (Type~3).
  \item Pure framing comparison (Type~4).
  \item General integration that does not hinge on conflict/verification (Type~6).
  \item Broad narrative or temporal questions (Types~8--9).
\end{itemize}

\textbf{Difficulty and caution:}
\begin{itemize}[leftmargin=1.2em, itemsep=2pt, topsep=2pt, parsep=0pt]
  \item Attempt this task only when you can clearly identify at least one explicit claim and at least one decisive cross-source evidence unit that tests it.
  \item Do not invent conflicts or numbers. If no strong conflict exists, formulate a precise non-support or inconsistency that is genuinely observable.
\end{itemize}

\textbf{Use of transcripts and frames:}
\begin{itemize}[leftmargin=1.2em, itemsep=2pt, topsep=2pt, parsep=0pt]
  \item Use transcripts to locate explicit claims and checkable statements.
  \item Use frames to extract on-screen numbers/labels/maps/scenes that refute or support these claims.
  \item The correct option must rely on at least one claim and at least one cross-source evidence unit in a clear contradictory or verifying relation.
\end{itemize}

\textbf{Structural requirements for the MCQ:}
\begin{itemize}[leftmargin=1.2em, itemsep=2pt, topsep=2pt, parsep=0pt]
  \item The stem must make it clear the task is to judge whether a claim holds up against other sources’ evidence.
  \item Each option (A--D) must combine two short sub-claims pairing a claim with evidence.
  \item \textbf{Exactly one} option must correctly describe the true evidence--claim relationship under joint evidence.
  \item Distractors may misidentify the contradicted claim, ignore crucial evidence, or combine non-conflicting items.
\end{itemize}

\end{promptbox}

\begin{promptbox}{Type 8: Cross-source Temporal Development}

\textbf{Goal.}
Generate \textbf{one hard multiple-choice question (MCQ)} that evaluates a solver’s ability to reconstruct the
\textbf{real-world chronological sequence of events} described across sources, and to integrate
which source contributes information about earlier, intermediate, and later phases.

\textbf{Core concept: Event-level temporal development.}
Type~8 concerns the timeline of the underlying incident, \textbf{not} the order in which sources published reports.
Focus on reconstructing the actual progression: initial trigger, escalation, response, and aftermath.

\textbf{What the question must focus on:}
\begin{itemize}[leftmargin=1.2em, itemsep=2pt, topsep=2pt, parsep=0pt]
  \item Ordering of real-world event phases across sources and how the full timeline emerges only after integration.
  \item Identifying which source provides information about earliest, intermediate, and latest phases.
\end{itemize}

\textbf{What the question must not reduce to:}
\begin{itemize}[leftmargin=1.2em, itemsep=2pt, topsep=2pt, parsep=0pt]
  \item Publication timing or who reported first.
  \item Non-temporal comparison (Type~2) or visual-only comparison (Type~3).
  \item Perspective differences (Type~4) or integration without ordering (Type~6).
  \item Narrative reconstruction without a temporal axis (Type~9) or summary (Type~10).
\end{itemize}

\textbf{Use of transcripts and frames:}
\begin{itemize}[leftmargin=1.2em, itemsep=2pt, topsep=2pt, parsep=0pt]
  \item Use transcripts for event-phase indicators (e.g., before/after references), sequential connectives, and consequences that presuppose earlier actions.
  \item Visuals may support phase cues (e.g., active emergency vs.\ recovery; nighttime vs.\ daytime), but rely on event-based sequencing rather than publication timing.
\end{itemize}

\textbf{Structural requirements for the MCQ:}
\begin{itemize}[leftmargin=1.2em, itemsep=2pt, topsep=2pt, parsep=0pt]
  \item The stem must specify that the task concerns \textbf{real-world temporal progression} across sources.
  \item Each option (A--D) must combine two short sub-claims describing event phases covered by different tags and their correct ordering.
  \item \textbf{Exactly one} option must match the chronological order implied across sources.
  \item Distractors may invert ordering, misassign phases, merge incompatible stages, or misread temporal cues.
\end{itemize}

\textbf{Source usage requirements:}
\begin{itemize}[leftmargin=1.2em, itemsep=2pt, topsep=2pt, parsep=0pt]
  \item Use \textbf{at least two distinct} source tags from the provided set.
  \item The question text must explicitly mention all chosen tags.
  \item The correct option must reflect event-level temporal progression.
  \item The support map lists exactly the tags whose descriptions contribute to the chronological chain.
\end{itemize}

\end{promptbox}

\begin{promptbox}{Type 9: Cross-source Narrative Reconstruction}

\textbf{Goal.}
Generate \textbf{one hard multiple-choice question (MCQ)} that requires reconstructing a \textbf{coherent comprehensive narrative}
by synthesizing complementary segments from multiple sources.

\textbf{Core concept: Narrative reconstruction.}
Type~9 focuses on the overall storyline that emerges only when sources are combined:
trigger/cause, escalation/spread, institutional or policy response, and aftermath or longer-term consequences.
Each source may cover only part of the story; the solver must integrate parts into one coherent narrative.

\textbf{What the question must focus on:}
\begin{itemize}[leftmargin=1.2em, itemsep=2pt, topsep=2pt, parsep=0pt]
  \item How different sources contribute different phases or components of the storyline.
  \item Logical and causal connections across phases, not isolated event matching.
  \item The item must require integrating multiple sources; a single-source narrative must be insufficient.
\end{itemize}

\textbf{What the question must not reduce to:}
\begin{itemize}[leftmargin=1.2em, itemsep=2pt, topsep=2pt, parsep=0pt]
  \item Publication ordering or a pure temporal-order task (Type~8).
  \item Pure details of event comparison without reconstructing the storyline (Type~2).
  \item Pure framing comparison (Type~4) or summary of agreement/disagreement (Type~10).
\end{itemize}

\textbf{Use of transcripts and frames:}
\begin{itemize}[leftmargin=1.2em, itemsep=2pt, topsep=2pt, parsep=0pt]
  \item Use transcripts to extract stages, explicit causal links, and details belonging to distinct narrative phases.
  \item Use frames to anchor stages visually (e.g., mid-crisis vs.\ cleanup) and to highlight distinct actors/locations across phases.
  \item The narrative must be grounded in provided evidence; do not invent events.
\end{itemize}

\textbf{Structural requirements for the MCQ:}
\begin{itemize}[leftmargin=1.2em, itemsep=2pt, topsep=2pt, parsep=0pt]
  \item The stem must indicate that the solver should reconstruct the combined storyline across sources.
  \item Each option (A--D) must combine two short sub-claims about narrative stages or links across sources.
  \item \textbf{Exactly one} option must describe a coherent, logically ordered narrative supported under cross-source integration.
  \item Distractors may scramble causal order, omit crucial phases, or misattribute stages, making the storyline inconsistent.
\end{itemize}

\end{promptbox}

\begin{promptbox}{Type 10: Multi-source Summary}

\textbf{Goal.}
Generate \textbf{one hard multiple-choice question (MCQ)} that evaluates whether a solver can form a \textbf{concise cross-source summary}
capturing both what sources agree on and where they differ.

\textbf{Core concept: Multi-source summary.}
Type~10 requires high-level synthesis across sources: shared core claims/event details and salient divergences in interpretations or implications.
The emphasis is on summarizing the joint coverage landscape, not any single report.

\textbf{What the question must focus on:}
\begin{itemize}[leftmargin=1.2em, itemsep=2pt, topsep=2pt, parsep=0pt]
  \item Stable elements across sources (core events/outcomes consistently reported; shared bottom-line assessment).
  \item Contested or varied elements (differences in main claims, interpretations, implications, or outlooks).
  \item The correct option should read like a compact meta-summary produced after comparing all included sources.
\end{itemize}

\textbf{What the question must not reduce to:}
\begin{itemize}[leftmargin=1.2em, itemsep=2pt, topsep=2pt, parsep=0pt]
  \item A narrow event details comparison (Type~2).
  \item Pure framing comparison only (Type~4).
  \item Narrative reconstruction (Type~9) or temporal ordering (Type~8).
  \item A single-source summary that ignores other sources.
\end{itemize}

\textbf{Use of transcripts and frames:}
\begin{itemize}[leftmargin=1.2em, itemsep=2pt, topsep=2pt, parsep=0pt]
  \item Use transcripts to identify key event details shared by all or most sources and major points of divergence.
  \item Do not invent consensus or conflict; base the summary strictly on provided material.
\end{itemize}

\textbf{Structural requirements for the MCQ:}
\begin{itemize}[leftmargin=1.2em, itemsep=2pt, topsep=2pt, parsep=0pt]
  \item The stem must indicate summarization \textbf{across} all chosen sources, capturing both consensus and disagreement.
  \item Each option (A--D) must combine two short clauses: one shared core element and one major divergence.
  \item \textbf{Exactly one} option must correctly capture the main shared event details/conclusions and the most salient differences.
  \item Distractors may overstate agreement, exaggerate conflicts, omit key shared points, or misrepresent what is disputed.
\end{itemize}

\textbf{Source usage requirements:}
\begin{itemize}[leftmargin=1.2em, itemsep=2pt, topsep=2pt, parsep=0pt]
  \item Use \textbf{at least two distinct} source tags from the provided set.
  \item The question text must explicitly mention all chosen tags.
  \item The correct option must summarize true consensus plus the most important disagreement.
  \item The support map lists the tags whose content underlies the correct summary (both shared and divergent parts).
\end{itemize}

\end{promptbox}

\subsection{Common rules for question generation}
\begin{promptbox}{Common rules}

\textbf{STRICT Requirements:}

\textbf{1) Sources (multi-source, size $\geq$ 2):}
\begin{itemize}[leftmargin=1.2em, itemsep=2pt, topsep=2pt, parsep=0pt]
  \item There are multiple source tags available (one per video/segment). Choose a subset of \textbf{distinct} tags with size $\geq$ 2.
  \item The question text must explicitly mention \textbf{all} chosen source tags somewhere
  (e.g., ``(TAG\_A vs TAG\_B vs TAG\_C)'' or ``reported by TAG\_A, TAG\_B and TAG\_C'').
  \item In \texttt{support\_map} for the correct option, list \textbf{all supporting tags} for that option
  (e.g., [``TAG\_A'', ``TAG\_B'', ``TAG\_C''] if all three consistently support the correct cross-source statement).
\end{itemize}

\textbf{2) Same story \& clear reasoning focus:}
\begin{itemize}[leftmargin=1.2em, itemsep=2pt, topsep=2pt, parsep=0pt]
  \item All chosen sources must discuss the \textbf{same} overarching news story (same underlying event or evolving storyline).
  \item You may combine multiple event dimensions (time, actor, magnitude, stance, visuals, place), but the reasoning must remain focused.
  \item If frames are present and the type involves visuals (Types 3, 5, 6):
  \begin{itemize}[leftmargin=1.2em, itemsep=1pt, topsep=1pt, parsep=0pt]
    \item Derive at least one decisive micro-cue from an image
    (e.g., lower-third text, ticker value, map/placename, logo, timestamp).
    \item Fall back to transcript-only cues only if no decisive visual cue exists.
  \end{itemize}
\end{itemize}

\textbf{3) Options (A--D):}
\begin{itemize}[leftmargin=1.2em, itemsep=2pt, topsep=2pt, parsep=0pt]
  \item Each option is a conjunction of two short event claims or separable sub-claims.
  \item \textbf{Exactly one} option is true when considering \textbf{all chosen sources together}.
  \item Each distractor must be locally plausible from at least one source but fails under full multi-source evidence.
  \item Visual claim requirement:
  \begin{itemize}[leftmargin=1.2em, itemsep=1pt, topsep=1pt, parsep=0pt]
    \item For Types 3, 5, and 6, at least two options must include a \textbf{visual} claim
    (e.g., on-screen numbers, labels, logos, maps, places).
  \end{itemize}
\end{itemize}

\textbf{4) Style / Safety:}
\begin{itemize}[leftmargin=1.2em, itemsep=2pt, topsep=2pt, parsep=0pt]
  \item Ban deictic pronouns: no ``one'', ``the other'', ``former'', ``latter'', ``first'', ``second''.
  \item No fabrication beyond transcripts or frames.
  \item When citing evidence, include at least one frame marker in the exact format:
  \texttt{[FRAME: TAG\_A \#1]} or \texttt{[FRAME: TAG\_B \#1]}.
  \item Evidence $\leq$ 160 characters; cite decisive micro-cues succinctly.
  \item The evidence must contain no more than two sentences.
\end{itemize}


\textbf{5) Output (strict JSON):}
\begin{lstlisting}[basicstyle=\rmfamily\small, aboveskip=2pt, belowskip=0pt]
{
  "layer":"hard",
  "type_id":"T1--T10",
  "question":"...",
  "options":{"A":"...","B":"...","C":"...","D":"..."},
  "correct_answer":"A|B|C|D",
  "evidence":"...",
  "support_map":{...},
  "evidence_frames":[...]
}
\end{lstlisting}

\end{promptbox}

\section{Human Verification}

\subsection{Verification Criteria}
\label{app:human_verification}
\begin{promptbox}{Human Verification Score Sheet}

\textbf{Correctness}

Evaluate whether the question and answer options are consistent with the associated video.
Check whether the gold answer aligns with human judgment based on the videos, and whether the question admits exactly one correct answer.

\begin{itemize}
    \item \textbf{pass}: The question and options align with the video evidence; the gold answer is clearly supported, and exactly one correct answer is unambiguous.
    \item \textbf{uncertainty}: The item is mostly consistent with the video, but evidence support is not fully decisive (e.g., minor ambiguities, borderline multiple-answer concern, or distractors not clearly ruled out).
    \item \textbf{fail}: The item is incorrect or inconsistent with the video; the gold answer is wrong or unsupported, no option is clearly correct, or multiple options could reasonably be considered correct.
\end{itemize}

\textbf{Naturalness}

Assign an overall judgment of how natural and human-written the entire item feels.
Consider wording quality, fluency, readability, and stylistic consistency across the stem and options.

\begin{itemize}
    \item \textbf{pass}: Fluent and natural overall; only minor issues that do not affect readability.
    \item \textbf{uncertainty}: Understandable but noticeably stiff, awkward, or uneven in style; would benefit from editing.
    \item \textbf{fail}: Very unnatural or difficult to read due to frequent grammar/phrasing issues; unacceptable without heavy rewriting.
\end{itemize}

\textbf{Task Compliance}

Evaluate whether the candidate multiple-choice question strictly matches the target type definition (core reasoning axis and explicit structural requirements).

\begin{itemize}
    \item \textbf{pass}: Clearly matches the target type definition and structural requirements, with no substantive off-type drift.
    \item \textbf{uncertainty}: Borderline match; partially aligned, but with minor-to-moderate slippage or missing/misaligned requirements.
    \item \textbf{fail}: Off-type or structurally invalid; the core reasoning axis does not match the type definition and the question would be categorized as a different type.
\end{itemize}

\end{promptbox}
\subsection{Inter-reviewer Consistency Metrics}\label{apd:iaa_metrics}

We quantify the consistency of human verification using
pairwise agreement $A_{\text{pair}}$.
Let $N$ denote the number of double-checked items and
$\ell_{ij}$ the label assigned by reviewer $j$ to item $i$
under a given verification dimension.
With three reviewers, the three unordered reviewer pairs
$(1,2)$, $(1,3)$, and $(2,3)$ are considered.
Pairwise agreement is defined as the proportion of exact
label matches across all items and reviewer pairs:
\begin{equation}
A_{\text{pair}}
=
\frac{1}{3N}
\sum_{i=1}^{N}
\big(
\mathbb{I}[\ell_{i1}=\ell_{i2}]
+
\mathbb{I}[\ell_{i1}=\ell_{i3}]
+
\mathbb{I}[\ell_{i2}=\ell_{i3}]
\big),
\end{equation}
where $\mathbb{I}[\cdot]$ is the indicator function.

\vspace{0.5em}
\noindent\textbf{Fleiss’~$\kappa$.}
In addition to pairwise agreement, we report Fleiss’~$\kappa$
to measure chance-corrected agreement among multiple reviewers.

Let $N$ denote the number of double-checked questions,
$R=3$ the number of reviewers,
and $K=3$ the number of possible labels
(\emph{pass}, \emph{uncertainty}, \emph{fail}).
Agreement is evaluated separately for each verification
dimension $d \in \mathcal{D}$, where
$\mathcal{D}=\{Correctness,Naturalness,TaskCompliance\}$.
For question $i$ and dimension $d$,
let $n_{i,d,k}$ denote the number of reviewers assigning
label $k$, with $\sum_{k=1}^{K} n_{i,d,k}=R$.

The per-question agreement for dimension $d$ is:
\begin{equation}
P_{i,d}
=
\frac{1}{R(R-1)}
\sum_{k=1}^{K}
n_{i,d,k}\bigl(n_{i,d,k}-1\bigr).
\end{equation}
The observed agreement is obtained by averaging over items:
\begin{equation}
\bar{P}_d
=
\frac{1}{N}
\sum_{i=1}^{N}
P_{i,d}.
\end{equation}
Let
\begin{equation}
p_{d,k}
=
\frac{1}{NR}
\sum_{i=1}^{N}
n_{i,d,k}
\end{equation}
denote the overall proportion of label $k$ for dimension $d$.
The expected agreement by chance is then:
\begin{equation}
P_{e,d}
=
\sum_{k=1}^{K}
p_{d,k}^2.
\end{equation}
Fleiss’~$\kappa$ for dimension $d$ is defined as:
\begin{equation}
\kappa_d
=
\frac{\bar{P}_d - P_{e,d}}{1 - P_{e,d}}.
\end{equation}

We finally obtain the Fleiss’~$\kappa$ of $0.63,0.57,0.66$ on correctness, naturalness, and task compliance three dimensions. Given the multi-class, multi-dimensional nature of our verification rubric and the highly imbalanced label distribution, dimension judgments across all
verification dimensions yields an overall
Fleiss’~$\kappa$ of $0.62$, indicating
\emph{moderate} agreement, which supports the reliability of the human
verification protocol.

\section{Frame Selection}\label{apd:frame_selection}

During the selection of frames, we first use GPT 5 to generate the captions from images, where we set 1 fps for every video. The prompt is:

\begin{promptbox}{Image Captioning}
\raggedright
\sloppy
\textbf{System Prompt:}
\begin{itemize}[leftmargin=1.2em, itemsep=2pt, topsep=2pt, parsep=0pt]
  \item You are a professional image annotator.
  \item Generate a dense and comprehensive caption for the image.
  \item Describe:
  \item 1. Every distinct object visible in the image
  \item 2. Each object's visual attributes (color, texture, material, approximate size)
  \item 3. Exact spatial layout and relative positioning between objects
  \item 4. Actions or interactions occurring in the scene
  \item 5. Background details and environmental context
   \item Be exhaustive but accurate.
  \item Avoid redundancy.
  \item Do not include interpretations or speculation.

  \item Output only the caption text.
\end{itemize}

\textbf{Human Prompt:}
\begin{itemize}[leftmargin=1.2em, itemsep=2pt, topsep=2pt, parsep=0pt]
  \item \textbf{Image:} <image>

\end{itemize}
\end{promptbox}

Then we equally split the video into 8 ranges, where we select the most related image for every range to form the final selected frames for further QA generation. The prompt is:

\begin{promptbox}{Image Caption Scoring}
\raggedright
\sloppy

\textbf{System Prompt:}
\begin{itemize}[leftmargin=1.2em, itemsep=2pt, topsep=2pt, parsep=0pt]
  \item You are an expert evaluator for multi-source, multi-modal news understanding.
  \item Your task is to evaluate an image caption based on how well it supports cross-source news comparison and integration.
  \item Assign a score from 0 to 10, where higher scores indicate greater usefulness for downstream multi-source reasoning types.
  \item Evaluate the caption according to the following criteria:
  \item 1. \textbf{Visual Specificity}: Does the caption precisely describe visible people, objects, actions, symbols, text, and setting?
  \item 2. \textbf{Evidence Explicitness}: Does the caption clearly present visual elements that can serve as concrete evidence for or against news claims?
  \item 3. \textbf{Cross-source Comparability}: Is the caption written in an observable manner that enables comparison across different news outlets?
  \item 4. \textbf{Narrative Neutrality}: Does the caption avoid subjective framing, emotional inference, or editorial interpretation?
  \item 5. \textbf{Temporal and Contextual Cues}: Does the caption include observable cues relevant to event timing, sequence, or stage of development?
  \item 6. \textbf{Multimodal Alignment Readiness}: Does the caption facilitate alignment with textual or audio evidence from news reports?
  \item Use the following scoring guidelines:
  \item \quad -- 0--2: Vague or generic, not useful for cross-source reasoning
  \item \quad -- 3--4: Basic description with limited analytical value
  \item \quad -- 5--6: Accurate but missing key details for comparison or integration
  \item \quad -- 7--8: Detailed, neutral, and supportive of multi-source analysis
  \item \quad -- 9--10: Highly detailed, evidence-centric, and ideal for cross-source, cross-modal news reasoning
  \item Output only the score and a brief justification.
\end{itemize}

\textbf{Human Prompt:}
\begin{itemize}[leftmargin=1.2em, itemsep=2pt, topsep=2pt, parsep=0pt]
  \item \textbf{Image Caption:} <caption text>
\end{itemize}

\end{promptbox}

\subsection{QA Evaluation}\label{sec:eval}
During evaluation of final results and MLLMs-as-judge difficulty filter, we implement the under MLLM evaluation. Here, $N$ is the number of videos in the news group, and $M$ is the number of frames evaluated. The prompt is:
\begin{promptbox}{Evaluation}
\raggedright
\sloppy
\textbf{System Prompt:}
\begin{itemize}[leftmargin=1.2em, itemsep=2pt, topsep=2pt, parsep=0pt]
  \item You are a careful multi-source QA solver.
  \item Answer the multiple-choice question based on the provided information.
  \item You must choose one choice from A, B, C, or D.
  \item Respond in STRICT JSON only. 
  \item Answer format: \{"answer":"A|B|C|D","confidence":0.0-1.0,"rationale":"short"\}
  
\end{itemize}

\textbf{Human Prompt:}
\begin{itemize}[leftmargin=1.2em, itemsep=2pt, topsep=2pt, parsep=0pt]

  \item \textbf{Question:}  <question>

  \item \textbf{Options:}  <option\_A><option\_B><option\_C><option\_D>

  \item \textbf{Transcripts:} \newline [TAG\_$1$]: <transcript\_$1$>\newline[TAG\_$2$]: <transcript $2$>\newline...\newline[TAG\_$N$]: <transcript\_$N$>

  \item \textbf{Videos:}  \newline[TAG\_$1$]:<Frame $1$><Frame\_$2$>...<Frame\_$M$>\newline[TAG\_$2$]:<Frame $1$><Frame\_$2$>...<Frame\_$M$>\newline...\newline[TAG\_$N$]:<Frame\_$1$><Frame\_$2$>...<Frame\_$M$>

\end{itemize}

\end{promptbox}

\subsection{Single-source Solvability Test}\label{apd:single-solve}

We want to make sure each QA is genuinely required multi-source reasoning. So we conduct the single-source solvability test. For each query, we randomly pick $i \in \{1,2,...,N\}$ video and its transcript. The prompt is:
\begin{promptbox}{Single-source Solvability}
\raggedright
\sloppy
\textbf{System Prompt:}
\begin{itemize}[leftmargin=1.2em, itemsep=2pt, topsep=2pt, parsep=0pt]
  \item You are a careful multi-source QA solver.
  \item Answer the multiple-choice question based on the provided information.
  \item You must choose one choice from A, B, C, or D.
  \item Respond in STRICT JSON only. 
  \item Answer format: \{"answer":"A|B|C|D","confidence":0.0-1.0,"rationale":"short"\}
  
\end{itemize}

\textbf{Human Prompt:}
\begin{itemize}[leftmargin=1.2em, itemsep=2pt, topsep=2pt, parsep=0pt]

  \item \textbf{Question:}  <question>

  \item \textbf{Options:}  <option\_A><option\_B><option\_C><option\_D>

  \item \textbf{Transcripts:} [TAG\_$i$]: <transcript\_$1$>

  \item \textbf{Videos:}  [TAG\_$i$]:<Frame $1$><Frame\_$2$>...<Frame\_$M$>

\end{itemize}

\end{promptbox}

\subsection{Ambiguity Analysis}\label{apd:ambiguity}
We further want to avoid the multiple-correct-choice QAs that generated from the GPT5, where we design an ambiguity analysis to mitigate them.  We denote for $p,q \in \{option\_A,option\_B,option\_C,option\_D\}, q\neq p$, where $p$ is the gold choice and $q$ is the pairwise choice that we want to analysis. For each QA, we need to check three times of ambiguity analysis to obtain the overall comparison with all other choices. The prompt is:

\begin{promptbox}{Ambiguity Analysis}
\raggedright
\sloppy
\textbf{System Prompt:}
\begin{itemize}[leftmargin=1.2em, itemsep=2pt, topsep=2pt, parsep=0pt]
  \item You are an expert in multiple-choice question quality control.
  \item You are given transcripts and images from multiple sources.
  \item A question has ONE designated correct answer (gold).
  \item You will compare ONE alternative option against the gold option,
judging whether the alternative could ALSO be considered correct.
  \item Severity is the level of problem with the alternative option:
  \item - 0.0-0.2: The alternative is clearly incorrect.
  \item - 0.3-0.5: The alternative is somewhat ambiguous but leans incorrect.
  \item - 0.6-0.9: The alternative is potentially correct or very ambiguous.
    \item - 1.0: The alternative is clearly correct, same as gold.
    \item Respond format:\{ "severity": 0.0-1.0\}
  
\end{itemize}

\textbf{Human Prompt:}
\begin{itemize}[leftmargin=1.2em, itemsep=2pt, topsep=2pt, parsep=0pt]

  \item \textbf{Question:}  <question>

  \item \textbf{Options:}  <option\_A><option\_B><option\_C><option\_D>
    \item \textbf{Gold option:}  <$p$>
    \item \textbf{Alternative:}  <$q$>
  \item \textbf{Transcripts:} \newline [TAG\_$1$]: <transcript\_$1$>\newline[TAG\_$2$]: <transcript $2$>\newline...\newline[TAG\_$N$]: <transcript\_$N$>

  \item \textbf{Videos:}  \newline[TAG\_$1$]:<Frame $1$><Frame\_$2$>...<Frame\_$M$>\newline[TAG\_$2$]:<Frame $1$><Frame\_$2$>...<Frame\_$M$>\newline...\newline[TAG\_$N$]:<Frame\_$1$><Frame\_$2$>...<Frame\_$M$>

\end{itemize}

\end{promptbox}

\section{Experiment Details}\label{apd:experiment_detail}
\subsection{Experiment Settings}
We deploy all the open-source models on our server, with 8 NVIDIA B200 GPUs, 400GB RAM, and Intel(R) Xeon(R) Platinum 8570 CPU. For all open-source models and close-source models, we set the max\_tokens at 4096, and the generation temperature at 0.0, and top\_p at 1.0. All frames  are resized such that the longer side was limited to 540 pixels, while the shorter side was scaled proportionally. The quality of the image is set to 65. During QA generation and QA quality check, we set the number of frames per video at 8. We set it to 4 during evaluation.

\subsection{Transcript Cleanness and Summarization}

First, we normalize raw ASR transcripts by unifying line breaks, collapsing redundant whitespace, and removing empty lines. Fragmented lines are then merged into a single coherent paragraph to eliminate artificial sentence breaks introduced by automatic speech recognition. Then, for those that exceed 1,000 characters, we summarize them to 1,000 characters each. Here, we use prompted LLM.  For each query, we use $N$ and $M$ to represent the number of videos and the number of frames in each video that we need to evaluate. We use $i \in \{1,2,...,N\}$ to present $i$-th video, where [TAG\_$i$] is the specific alias of it. 

\begin{promptbox}{Transcript Summarization}
\raggedright
\sloppy
\textbf{System Prompt:}
\begin{itemize}[leftmargin=1.2em, itemsep=2pt, topsep=2pt, parsep=0pt]
  \item You are a professional news transcript summarizer.
  \item Summarize the transcript faithfully.
  \item - Preserve key entities, events, dates, and numbers
  \item - Do NOT add interpretation or speculation
  \item - Max length: 1000 characters
  \item Output plain text only.
\end{itemize}

\textbf{Human Prompt:}
\begin{itemize}[leftmargin=1.2em, itemsep=2pt, topsep=2pt, parsep=0pt]
  \item \textbf{Transcripts:} [TAG\_$i$]: <transcript\_$i$>

\end{itemize}
\end{promptbox}


%% file: tables_figures_tex/data_statistics.tex
\begin{table*}[h]
\centering
\small
\setlength{\tabcolsep}{4pt}
\renewcommand{\arraystretch}{1.15}
\caption{\textbf{Distribution of News Groups, Videos, and Question Types across Domains in VNU-Bench.}
For each domain, we report the number of news groups, total videos, the counts of questions for each task type (T1--T10), and the total number of questions.}
\begin{tabular}{l |c c |c c c c c c c c c c| c}
\hline
\textbf{Domain} 
& \textbf{News Groups} 
& \textbf{Videos} 
& \textbf{T1} & \textbf{T2} & \textbf{T3} & \textbf{T4} & \textbf{T5} 
& \textbf{T6} & \textbf{T7} & \textbf{T8} & \textbf{T9} & \textbf{T10} 
& \textbf{Total QA} \\
\hline
Business \& Economy        & 98 & 326 & 64 & 64 & 48 & 48 & 72 & 53 & 51 & 44 & 62 & 60 & 566 \\
Climate \& Environment    & 67 & 215 & 38 & 26 & 43 & 38 & 34 & 46 & 28 & 33 & 22 & 31 & 339 \\
Culture \& Arts           & 48 & 147 & 62 & 59 & 37 & 41 & 39 & 42 & 49 & 46 & 47 & 43 & 465 \\
Health                    & 68 & 235 & 31 & 26 & 46 & 35 & 36 & 30 & 27 & 28 & 25 & 22 & 306 \\
Science \& Technology     & 64 & 189 & 38 & 45 & 39 & 40 & 42 & 43 & 44 & 18 & 36 & 39 & 384 \\
US \& Canada               & 50 & 181 & 23 & 30 & 38 & 25 & 26 & 17 & 24 & 28 & 23 & 20 & 254 \\
World                     & 34 & 112 & 15 & 21 & 26 & 21 & 28 & 19 & 12 & 16 & 12 & 17 & 187 \\
\hline
\textbf{Total}            
& 429 & 1,405
& 271 & 271 & 277 & 248 & 277 & 250 & 235 & 213 & 227 & 232 
& 2,501 \\
\hline
\end{tabular}
\label{tab:data_distribution}
\end{table*}